\documentclass{article}
\usepackage{arxiv}
\usepackage[utf8]{inputenc}
\usepackage[T1]{fontenc}
\usepackage{amsmath,amssymb,amsfonts}
\usepackage{graphicx}
\usepackage{algorithm}
\usepackage{url}
\usepackage{microtype}
\usepackage{textcomp}
\usepackage[table]{xcolor}
\usepackage{wrapfig}
\usepackage{booktabs}
\usepackage[numbers,sort&compress]{natbib}
\usepackage{algpseudocode}

\usepackage{amsmath,amsfonts,bm}

\def\eqref#1{equation~\ref{#1}}
\def\1{\bm{1}}

\DeclareMathAlphabet{\mathsfit}{\encodingdefault}{\sfdefault}{m}{sl}
\SetMathAlphabet{\mathsfit}{bold}{\encodingdefault}{\sfdefault}{bx}{n}

\usepackage[hidelinks]{hyperref}

\def\BibTeX{{\rm B\kern-.05em{\sc i\kern-.025em b}\kern-.08em
    T\kern-.1667em\lower.7ex\hbox{E}\kern-.125emX}}

\begin{document}
\title{Optimization Encoders:
Rethinking Second-Order Meta-Learning for Neural Fields}

\author{%
\textbf{Rudolf L. M. van Herten}$^{1,2}$%
\thanks{Contributed equally}%
\thanks{Corresponding author: \texttt{rlv4001@med.cornell.edu}}
\qquad
\textbf{Soufiane Ben Haddou}$^{3}$\footnotemark[1]
\qquad
\textbf{Rachit Saluja}$^{2}$
\qquad
\textbf{Johannes C. Paetzold}$^{1,2}$ \\[0.6em]
$^{1}$Weill Cornell Medicine
\qquad
$^{2}$Cornell Tech
\qquad
$^{3}$Amsterdam UMC
}

\maketitle

\begin{abstract}
Conditional neural fields represent signals continuously, but their effectiveness depends on how the conditional latent representations are inferred from observed data. In meta-learning, this encoding occurs through gradient updates induced by the decoder, tying representation learning directly to decoder design. We formalize this connection by interpreting latent optimization as an optimization encoder, unifying the roles of second-order differentiation, latent parameterization, and task supervision. This concept enables second-order meta-learning for end-to-end training of the encoding procedure alongside the decoder, and clarifies which learning pathway first-order approximations discard. Guided by this view, we introduce \emph{Attentive Latent Fields} (MetaLF), an equivariant transformer-based neural field that contextualizes a latent pointcloud through self-attention. These interactions shape both field predictions and the updates that construct their representation, allowing local observations to inform coherent non-local structure. Disentangling the inner encoding objective from outer task supervision unifies reconstruction, classification, and segmentation within an end-to-end meta-learning framework, using reconstruction-only latent adaptation at test time. Controlled experiments on polynomial fields link latent coordination to lower effective rank and stronger alignment with the underlying function space. Across image and 3D shape reconstruction, MetaLF improves fidelity within three to five gradient updates, while supporting semantic prediction across images, shapes, and volumes. Together, these findings position the optimization encoder perspective as a unified basis for designing neural fields around how representations are constructed, coordinated, and used. Code is made publicly available on \href{https://github.com/RoelvH97/MetaLF}{GitHub}.
\end{abstract}

\section{Introduction}

In recent years, neural fields~\citep{xie2022neural} have gained traction for their ability to represent discretely sampled data as continuous signals. A neural field parameterizes a function that maps coordinates to field values, i.e. \(f_\theta:\mathbb{R}^{d_{\mathrm{in}}}\rightarrow\mathbb{R}^{d_{\mathrm{out}}}\). For example, an RGB image can be represented by \(f_\theta:\mathbb{R}^2\rightarrow\mathbb{R}^3\), which maps each spatial coordinate to its corresponding color. Conditional neural fields (CNFs) extend this formulation by conditioning the coordinate-based function on a sample-specific latent representation \(z_i\in\mathcal{Z}\)~\citep{park2019deepsdf}:
\[
f_\theta:
\mathbb{R}^{d_{\mathrm{in}}}
\times
\mathcal{Z}
\rightarrow
\mathbb{R}^{d_{\mathrm{out}}}.
\]
This allows the shared parameters \(\theta\) to capture structure across a dataset, while \(z_i\) encodes variation specific to sample \(i\).

\begin{figure*}[t]
\centering
     \includegraphics[width=\textwidth,trim={0.6cm 0.8cm 0.6cm 0.5cm}]{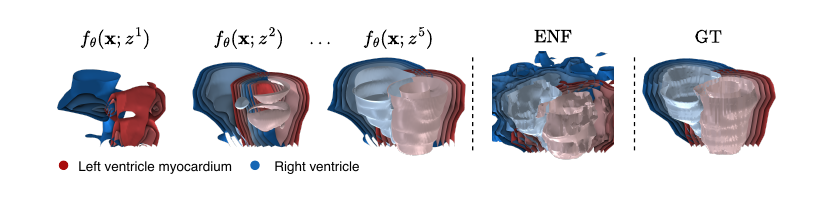}
     \caption{
     Example use case of MetaLF meta-learning latent cardiac shape representations~\citep{bernard2018deep}. The first three panels illustrate smooth shape recovery as a shared latent pointcloud~$z$ is refined over $k = 1, \ldots, 5$ gradient steps. MetaLF learns contextualized latent representations that preserve detail while capturing the underlying function, an ill-posed task for local representation models such as ENF~\citep{wessels2025grounding}.
     }
     \label{fig:example}
\end{figure*}

Learning a CNF requires a mechanism for obtaining sample-specific representations. Existing approaches use amortized encoders \citep{zhang20233dshape2vecset}, auto-decoding \citep{park2019deepsdf}, or meta-learning \citep{finn2017model,nichol2018first,sitzmann2020metasdf}. In the \textit{functa} framework \citep{dupont2022data}, a representation \(z_i^K\) is constructed from a shared initialization \(z^0\) through a small, fixed number of gradient updates. The inner loop adapts the latent representation while keeping the decoder fixed; the outer loop then updates the shared parameters to improve reconstruction after latent adaptation. This enables new signals to be represented within only a few gradient steps.  Figure~\ref{fig:example} illustrates this few-step fitting process for cardiac shape reconstruction.

Second-order meta-learning is central to the effectiveness of this approach, with first-order approximations typically yielding substantially worse reconstructions. Yet treating this distinction primarily as a choice of optimization method obscures its implications for model design. Second-order differentiation propagates the outer objective through the unrolled latent updates, accounting for how the decoder influences the representation obtained during adaptation. First-order variants omit this dependence on the decoder parameters. To understand what this gradient pathway learns, we revisit meta-learned CNFs from an encoder--decoder perspective.

\begin{figure*}[t]
\centering
     \includegraphics[width=\textwidth,trim={0.4cm 1.0cm 0.4cm 0.5cm}]{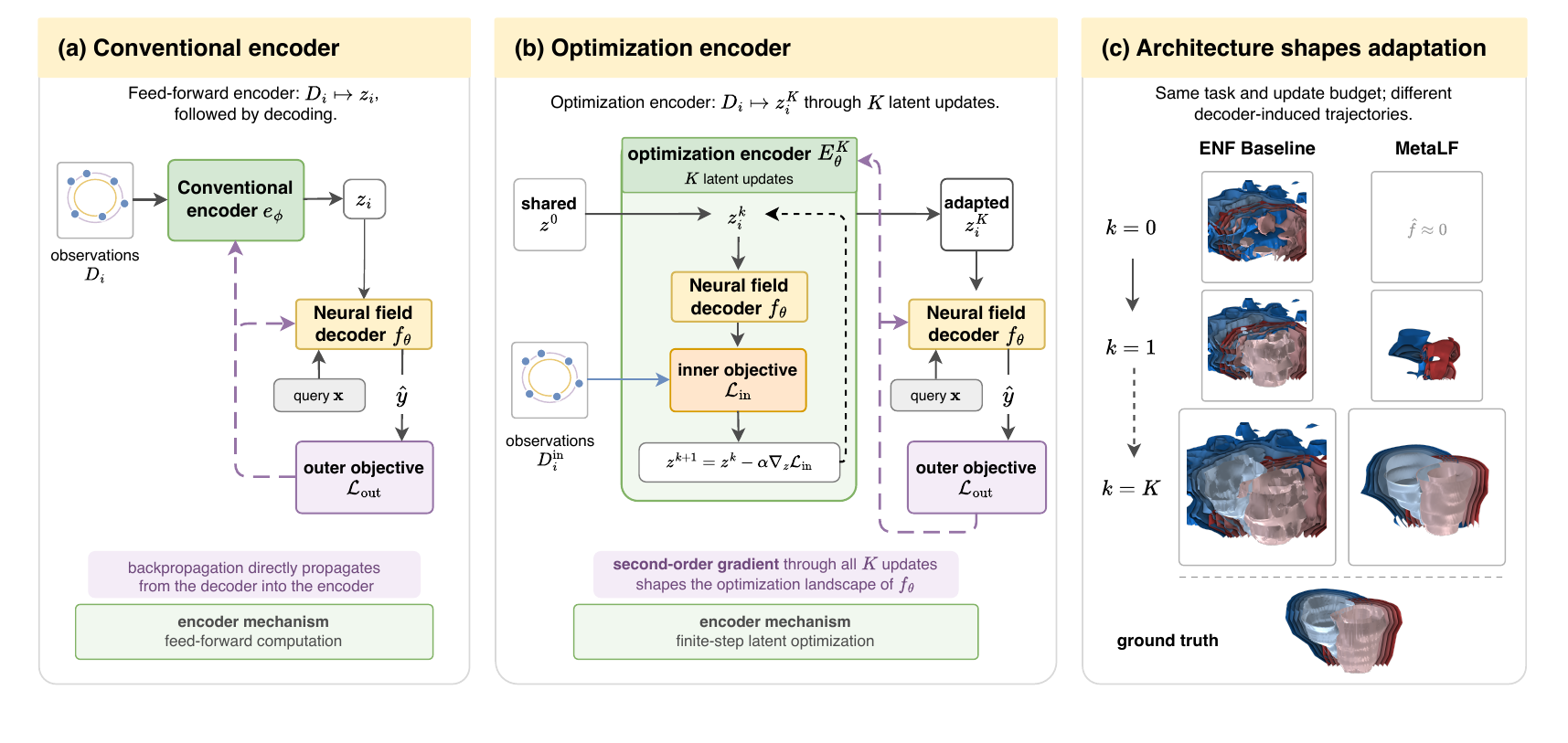}
     \caption{
     \textbf{Second-order meta-learning trains an optimization encoder whose behaviour is shaped by decoder architecture.} (a) A conventional encoder maps an observed signal to a latent representation \(z_i\), which the decoder evaluates at a query coordinate \(\mathbf{x}\). (b)~In meta-learned neural fields, an explicit encoder is replaced by \(K\) gradient updates on \(z\), forming an optimization encoder. Second-order differentiation propagates the outer objective \(\mathcal{L}_{\mathrm{out}}\) through these unrolled updates, jointly shaping the finite-step encoding behavior and the decoder \(f_\theta\). (c)~Because the decoder induces the latent updates, its architecture shapes the resulting adaptation trajectory. ENF represents a local fitting model, lacking direct exchange of context among localized latents. MetaLF instead coordinates these latents through attention. This shared context produces coherent updates that quickly capture non-local structure, as illustrated by fitting cardiac signed distance functions.
     }
     \label{fig:motivation}
\end{figure*}

Building on the view of optimization as encoding \citep{xu2020metafun}, we interpret the inner loop as an \emph{optimization encoder}: a mapping from observed data to a latent representation, implemented through gradient updates rather than a separate feed-forward network (Figure~\ref{fig:motivation}a-b). Its encoding behavior is determined by the decoder's backward pass, together with the latent parameterization and inner objective. Second-order meta-learning therefore trains the encoding procedure and decoder end-to-end, just as a conventional encoder--decoder model learns both how to construct a representation and how to use it. This interpretation makes the decoder architecture a design choice for encoding itself: its expressivity and interactions shape both field predictions and the updates through which their representation is inferred~(Figure~\ref{fig:motivation}c).

This observation motivates meta-learned \emph{Attentive Latent Fields} (MetaLF), an equivariant transformer-based neural field that combines spatially localized representations with contextual interactions. Equivariant neural fields (ENFs) \citep{wessels2025grounding} provide a strong foundation for local fitting by conditioning predictions on a latent pointcloud, but lack direct information exchange among latent elements before decoding. Drawing on attention-based shape autoencoders \citep{zhang20233dshape2vecset}, MetaLF extends ENF with self-attention that contextualizes the content of latents. These interactions shape both decoding and the gradients used for encoding, allowing local observations to inform coherent non-local structure while retaining fine detail.

\begin{figure*}[t]
\centering
     \includegraphics[width=\textwidth,trim={0.45cm 0.4cm 0.45cm 0.3cm}]{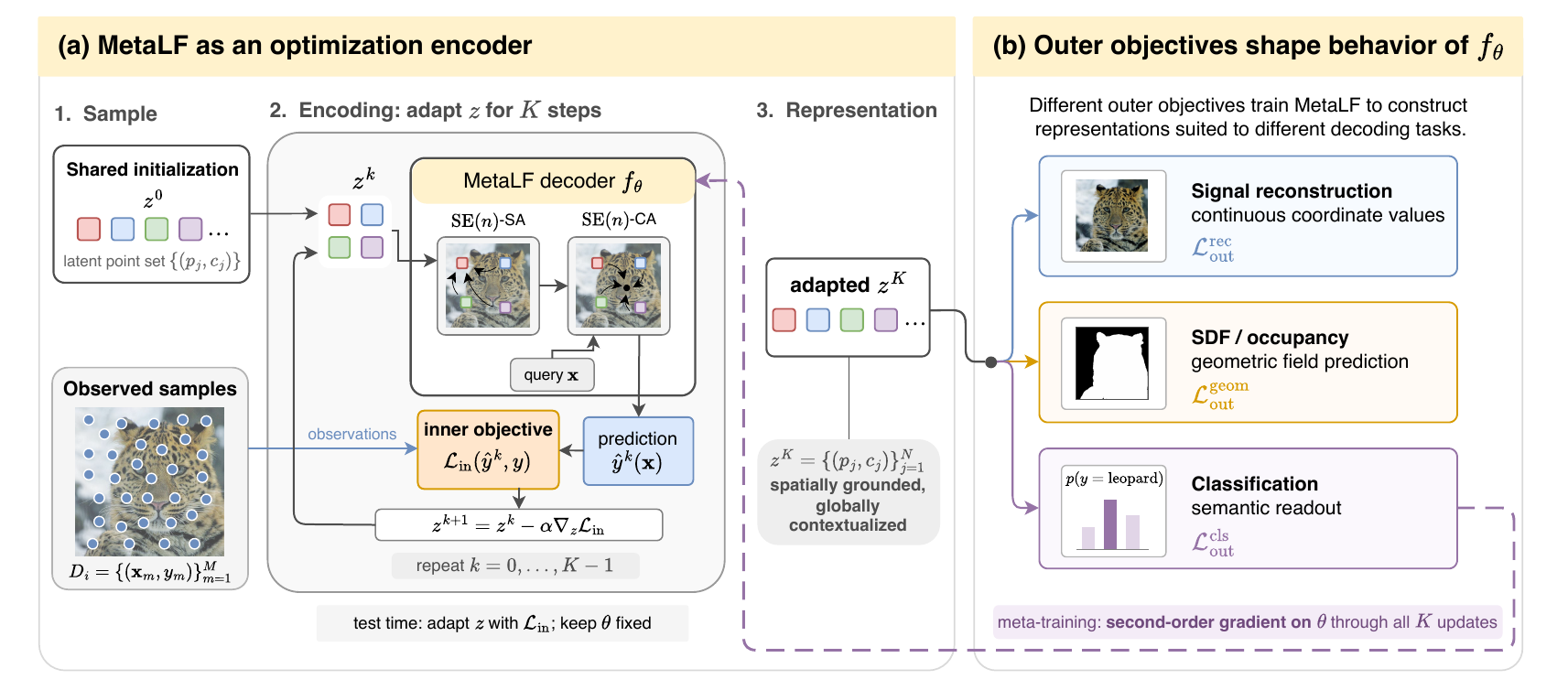}
     \caption{MetaLF as an optimization encoder. (a) \(K\) gradient updates encode observations into a spatially grounded, globally contextualized latent representation through \(\mathrm{SE}(n)\)-equivariant self-attention (SA) layers and cross-attention (CA) with a query coordinate. (b) During meta-training, task-specific outer objectives can shape the decoder through second-order gradients across the adaptation steps. Global contextualization of the latent point set therefore enables flexible meta-training: the same inner-loop encoder can learn representations tailored to diverse downstream objectives.}
     \label{fig:main}
\end{figure*}

The encoder--decoder perspective also clarifies the distinct roles of the inner and outer objectives. The inner objective determines how observations construct a representation, while the outer objective determines what that representation should support, i.e. they need not describe the same task. With end-to-end differentiation through adaptation, reconstruction updates can learn to produce representations useful for classification or segmentation, even though semantic labels are available only during meta-training. MetaLF accommodates these objectives through task-specific outputs while retaining reconstruction-only adaptation at test time (see Figure~\ref{fig:main}).

Our contributions follow from this perspective:
\begin{enumerate}
    \item We formalize few-step meta-learning as an optimization encoder, identifying second-order differentiation as the pathway through which the outer objective trains the encoding procedure alongside the decoder. This clarifies which learning pathway first-order approximations discard.

    \item Because the decoder also determines how representations are inferred, we introduce \mbox{MetaLF} to coordinate a latent pointcloud through equivariant self-attention. On controlled polynomial fields, this coordination yields more compact, function-aligned representations and improves reconstruction, linking architectural design to encoding behavior.

    \item Extending this design to images and 3D shapes, we demonstrate improved reconstruction over \textit{functa}-based baselines within three to five gradient updates, showing that contextual latent interactions benefit few-step fitting beyond the controlled setting.

    \item Finally, our optimization encoder perspective disentangles the roles of the inner and outer loops, allowing distinct encoding and decoding objectives, respectively. We demonstrate end-to-end learning of segmentation and classification through reconstruction-based inner-loop encoding and outer-loop semantic supervision on images, shapes, and volumes.
\end{enumerate}
\section{Background}
\label{sec:background}

\paragraph{Equivariant neural fields}
\label{sec:background_enfs}

ENFs ground their conditioning variables in geometry~\citep{wessels2025grounding}: a signal is represented by a latent point set \(z=\{(p_j,c_j)\}_{j=1}^{N}\), where \(p_j\in G\) is a pose in a transformation group \(G\) and \(c_j\in\mathbb{R}^{c}\) is an appearance feature. A group element \(g\in G\) acts on this representation by transforming its poses, \(gz=\{(gp_j,c_j)\}_{j=1}^{N}\)~\citep{cohen2019general}. The corresponding field satisfies the steerability property
\[
f_\theta(g^{-1}\mathbf{x};z)
=
f_\theta(\mathbf{x};gz),
\qquad \forall g\in G,
\]
such that transformations of the field correspond to transformations of the latent point set. ENFs implement this property by expressing interactions between query coordinates and latent elements through relative geometric quantities, specifically accounting for the group of roto-translations~$\mathrm{SE}(n)$. The same local mapping is therefore shared across latent elements and transformed signals. This weight sharing provides an inductive bias for learning localized representations in which each latent element describes information associated with a region of the signal.

\paragraph{Gradient-based latent meta-learning}
\label{sec:background_meta}

Let the observations for sample \(i\) be divided into an inner set and an outer set. Starting from a shared initialization \(z^0\), the inner loop applies \(K\) gradient updates to the latent representation using an objective \(\mathcal{L}_{\mathrm{in}}\), producing \(z_i^K\). The decoder parameters \(\theta\) remain fixed during this adaptation. The outer loop then updates \(\theta\) according to an objective \(\mathcal{L}_{\mathrm{out}}\) evaluated after adaptation, thus corresponding to an instance of CAVIA~\citep{zintgraf2019fast}. The two objectives may use different observations or losses.

Since every latent update is computed through \(f_\theta\), the adapted representation \(z_i^K\) depends on the decoder parameters. Model-agnostic meta-learning (MAML) accounts for this dependency by differentiating the outer objective through the sequence of inner updates~\citep{finn2017model}. The resulting meta-gradient contains mixed second-order derivatives with respect to \(z_i\) and \(\theta\). First-order meta-learning (FOMAML) omits these terms to reduce the computational cost of meta-training~\mbox{\citep{nichol2018first}}. In this scenario, the decoder still receives a direct gradient from \(\mathcal{L}_{\mathrm{out}}\), and \(\theta\) still shapes every subsequent inner update, but it omits the pathway through which \(\mathcal{L}_{\mathrm{out}}\) can steer the influence. Details on the MAML and FOMAML training loops are provided in Appendix~\ref{app:meta_learning}.
\section{Meta-learning as an optimization encoder}
\label{sec:optimization_encoder}

\subsection{Unrolling latent adaptation}
\label{sec:unrolling}

Let \(\mathcal{D}_i\) denote the coordinate-value observations available for sample \(i\). Starting from a shared initialization \(z^0\), the inner loop adapts the latent representation by minimizing \(\mathcal{L}_{\mathrm{in}}\), while keeping the decoder parameters \(\theta\) fixed. We write one gradient update as \(U_{\theta,\mathcal{D}_i}\) and define the complete adaptation procedure by
\begin{equation}
\label{eq:optimization-encoder}
\begin{aligned}
U_{\theta,\mathcal{D}_i}(z)
&=
z-\alpha\nabla_z
\mathcal{L}_{\mathrm{in}}(\theta,z;\mathcal{D}_i),
&
E_\theta^K(\mathcal{D}_i)
&:=
U_{\theta,\mathcal{D}_i}^{\circ K}(z^0)
=
z_i^K ,
\end{aligned}
\end{equation}
where \(U^{\circ K}\) denotes \(K\) repeated applications of the update operator. The map \(E_\theta^K:\mathcal{D}_i\mapsto z_i^K\) is the \emph{optimization encoder}, whose output is decoded at a query coordinate \(\mathbf{x}\) as \(f_\theta(\mathbf{x};z_i^K)\).

This computation has the same input-output structure as a conventional encoder-decoder: observations are mapped to a latent representation, which is subsequently used to produce predictions. The difference lies in the implementation of the encoder. Rather than applying a separate feed-forward network, \(E_\theta^K\) constructs its representation through a sequence of gradient updates with shared dynamics.

The optimization encoder is determined by the initialization \(z^0\), the latent parameterization, the inner objective, the learning rate \(\alpha\), the number of updates \(K\), and the decoder \(f_\theta\). In particular, it depends on \(\theta\) because every latent update is computed through the decoder. The resulting formulation is a finite-step bilevel optimization problem, in which the outer objective depends on the trajectory induced by the inner updates~\citep{maclaurin2015gradient,franceschi2018bilevel}.

\subsection{Training through the optimization encoder}
\label{sec:optimization-encoder-gradient}

Since \(E_\theta^K\) is parameterized entirely by \(\theta\), the outer objective can only train it through the decoder. Let
\(\mathcal{L}_{\mathrm{out}}^i(\theta,z_i^K)\) denote the outer loss for sample \(i\), and define\mbox{
\(
P_i^k := \partial z_i^k / \partial \theta.
\)}
Differentiating the inner update in Eq.~\ref{eq:optimization-encoder} gives
\[
P_i^{k+1}
=
A_i^k P_i^k-\alpha B_i^k,
\qquad
A_i^k
=
I-\alpha\nabla_{zz}^2
\mathcal{L}_{\mathrm{in}}^i(\theta,z_i^k),
\qquad
B_i^k
=
\nabla_{z\theta}^2
\mathcal{L}_{\mathrm{in}}^i(\theta,z_i^k).
\]
Thus, even though \(\theta\) is fixed within the inner loop, it affects every update through the gradient field induced by the decoder. Since \(z^0\) is a separate parameter rather than a function of \(\theta\), \(P_i^0= \frac{\partial z^0}{\partial \theta} =0\) whether \(z^0\) is fixed or learned, and the exact meta-gradient decomposes as
\begin{equation}
\label{eq:meta-gradient-decomposition}
\nabla_\theta
\mathcal{L}_{\mathrm{out}}^i(\theta,z_i^K)
=
\underbrace{
\partial_\theta
\mathcal{L}_{\mathrm{out}}^i(\theta,z_i^K)
}_{\text{direct decoder gradient}}
+
\underbrace{
(P_i^K)^\top
\nabla_z
\mathcal{L}_{\mathrm{out}}^i(\theta,z_i^K)
}_{\text{gradient through the optimization encoder}} .
\end{equation}
The first term updates the decoder while holding the inferred representation fixed. The second differentiates through the computation that produced \(z_i^K\). As shown in Appendix~\ref{app:expanded_meta_gradient}, this term contains mixed derivatives \(B_i^k\), which inject the dependence on \(\theta\) at each inner step, and products of \(A_i^k\), which transport these contributions through the remaining updates.

This yields a direct correspondence with conventional encoder--decoder training. For an amortized encoder \(e_\phi\), the outer objective propagates to the encoder according to
\[
\nabla_\phi
\mathcal{L}_{\mathrm{out}}^i
\bigl(\theta,e_\phi(\mathcal{D}_i)\bigr)
=
\left(
\frac{\partial e_\phi(\mathcal{D}_i)}{\partial\phi}
\right)^\top
\nabla_z\mathcal{L}_{\mathrm{out}}^i.
\]
Full MAML preserves the analogous path through
\(E_\theta^K\): the outer loss trains not only how the decoder uses a representation, but also how its inner gradient dynamics construct that representation. A first-order approximation removes the second derivatives of the inner loss, replacing \(A_i^k\) by \(I\) and \(B_i^k\) by \(0\). Since \(P_i^0=0\), this gives \(P_i^K\approx0\), leaving only the direct term in Eq.~\ref{eq:meta-gradient-decomposition}.

The resulting distinction is therefore structural: Conventional encoder-decoder training backpropagates through a feed-forward encoder, MAML backpropagates through an optimization encoder, and first-order training truncates the encoder path. In this view, second-order differentiation is the mechanism by which the outer objective trains the finite-step inference procedure itself.

\subsection{Inner and outer objectives}

CNFs are typically optimized as a variation of autoencoders, where $\mathcal{L}_{in}$ and $\mathcal{L}_{out}$ are the same reconstruction term. More generally, the inner and outer objectives may use different data or supervision within the bilevel formulation of meta-learning~\citep{finn2017model,franceschi2018bilevel}. Let \(D_i^{\mathrm{in}}\) denote the observations used for adaptation and \(T_i^{\mathrm{out}}\) the supervision available during meta-training. The general formulation is
\begin{equation}
\begin{aligned}
    z_i^{k+1}
        &= z_i^k-\alpha\nabla_z
        \mathcal{L}_{\mathrm{in}}(\theta,z_i^k;D_i^{\mathrm{in}}),
        \qquad z_i^0=z^0, \\
    \theta^*
        &= \arg\min_{\theta}\sum_i
        \mathcal{L}_{\mathrm{out}}(\theta,z_i^K;T_i^{\mathrm{out}}).
\end{aligned}
    \label{eq:flexible_objectives}
\end{equation}
The inner objective determines how observations are encoded into \(z_i^K\), whereas the outer objective specifies what the representation should support. The inner objective must be computable from test-time observations. However, the outer objective may use supervision available only during meta-training.

Full second-order differentiation couples the objectives through the second term in Eq.~\ref{eq:meta-gradient-decomposition}. The gradient of $\mathcal{L}_{\mathrm{out}}$ with respect to the adapted representation is propagated through the updates induced by $\mathcal{L}_{\mathrm{in}}$. The outer objective can therefore train how the inner objective constructs a representation, even when the two losses express different tasks.

\section{Attentive Latent Fields}

Most CNFs are designed from a neural-field-first perspective: sample-specific variation is captured through a latent representation \(z\), while architectural design focuses primarily on mapping \(z\) and a query coordinate to a field value. In meta-learned neural fields, interactions among latent elements can shape both field decoding and the gradient updates through which \(z\) is inferred. \mbox{ENFs~\citep{wessels2025grounding}} take an important step by providing spatial structure to \(z\), but do not explicitly model interactions among its latent elements before field decoding. The optimization encoder perspective suggests that such interactions are consequential not only for decoding, but also for the gradient dynamics through which \(z\) is constructed. We therefore propose MetaLF, a decoder-only transformer~\citep{vaswani2017attention} that uses equivariant self-attention to shape and coordinate a latent point set before evaluating the output field at query coordinates.

\begin{wrapfigure}[19]{l}{0.5\textwidth}
     \centering
     \includegraphics[width=0.5\textwidth, trim=0.7cm 0.6cm 0.7cm 0.6cm]{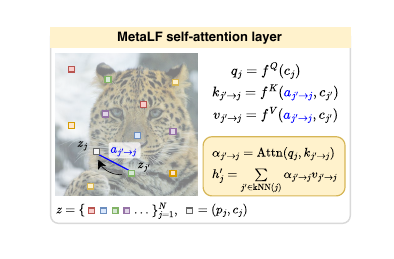}
     \caption{A receiving latent \(j\) attends to neighbouring source latents \(j'\) using bi-invariant geometric attributes \(a_{j'\to j}\). Keys and values combine this geometry with the source content \(c_{j'}\), and their attention-weighted aggregation produces the updated feature \(h'_j\).}
     \label{fig:self_attention}
\end{wrapfigure}

\paragraph{Architecture}
Let
\(
    z=\{(p_j,c_j)\}_{j=1}^{N}
\)
be a latent pointcloud, where \(p_j\) is the pose of latent \(j\) and \(c_j\) is its content. Both are sample-specific variables adapted by the optimization encoder. Inspired by transformer-based shape autoencoders~\citep{zhang20233dshape2vecset}, MetaLF projects the latent contents to token features \(h_j^0=W_c c_j\) and processes them with $L=N_\mathrm{SA}$ layers of $\mathrm{SE}(n)$-equivariant self-attention as illustrated in Figure~\ref{fig:self_attention}.

Self-attention (SA) is leveraged to update the content features for each latent point based on its latent neighbourhood, while poses are used as the geometric reference frame for attention. Each SA-block applies layer-normalized (LN) spatial self-attention followed by a GeGLU feed-forward network (FFN), with residual connections around both operations:
\[
\begin{aligned}
    \widetilde h^\ell
        &=h^\ell+
          \operatorname{SA}_{\ell}
          \bigl(\operatorname{LN}(h^\ell),p\bigr),\\
    h^{\ell+1}
        &=\widetilde h^\ell+
          \operatorname{FFN}_{\ell}
          \bigl(\operatorname{LN}(\widetilde h^\ell)\bigr).
\end{aligned}
\]
After the SA stack, the contextualized latent pointcloud
\(\{(p_j,h_j^L)\}_{j=1}^{N}\) is interpolated at a query coordinate using ENF cross-attention. The purpose of self-attention is therefore to separate communication among latent variables from field decoding: latent information is first combined and distilled, and only then mapped back to the output domain.

\paragraph{Equivariant attention and decoding}

We extend ENF cross-attention to self-attention by expressing spatial interactions through bi-invariant relative attributes. Following \citep{wessels2025grounding}, we denote bi-invariant geometric attributes in \textcolor{blue}{blue}:
\mbox{\(
    \textcolor{blue}{a_{j'\to j}=a(p_j,p_{j'})}
\)}
where
\[
    \textcolor{blue}{a_{j'\to j}} =
    \begin{cases}
        p_j-p_{j'},
        & G=\mathbb{R}^n,\\[2mm]
        \left(R_{j'}^\top(t_j-t_{j'}),\,R_{j'}^\top R_j\right),
        & G=\mathrm{SE}(n),
    \end{cases}
\]
with \(p_j\in\mathbb{R}^n\) for translation equivariance and
\(p_j=(t_j,R_j)\in\mathrm{SE}(n)\) in the roto-translation setting. Unlike ENF cross-attention, each self-attention query corresponds to a latent with its own content. We therefore compute queries from the receiving latent,
\[
    q_j=W_q\bar h_j,\qquad \bar h_j=\operatorname{LN}(h_j),
\]
and use relative geometry as the base of a FiLM-modulated key~\citep{perez2018film}:
\[
\begin{aligned}
    \textcolor{blue}{r_{j'\to j}}
        &= \phi(\textcolor{blue}{a_{j'\to j}}), \qquad
    u_{j'} = W_k\bar h_{j'},\qquad
    k_{j'\to j}
        = \textcolor{blue}{r_{j'\to j}}
        \odot\left(1+\gamma_k(u_{j'})\right)+\beta_k(u_{j'}),
\end{aligned}
\]
where \(\phi\) is a Gaussian random Fourier feature
embedding~\citep{tancik2020fourier}. This parameterization gives attention a geometric base while allowing the source content to modify the interaction. The resulting attention logits are
\[
    \ell_{j'\to j}
    =
    \frac{q_j^\top k_{j'\to j}}{\sqrt{d_h}}
    -
    \frac{\lVert
        \textcolor{blue}{a_{j'\to j}^{\mathrm{pos}}}
    \rVert_2^2}{\sigma^2},
\]
where the second term is the Gaussian spatial windowing proposed by~\cite{wessels2025grounding} in order to promote locality. Values are similarly conditioned on relative geometry:
\[
    v_{j'\to j}
    =
    W_v\bar h_{j'}\odot
    \left(1+\gamma_v(\textcolor{blue}{r_{j'\to j}})\right)
    +\beta_v(\textcolor{blue}{r_{j'\to j}}).
\]
Attention aggregates these values over the \(K_{\mathrm{SA}}\) nearest latent poses. Together with the Gaussian window, this promotes local feature interaction, while stacking several blocks allows information to propagate over longer distances.

MetaLF subsequently applies ENF-style cross-attention between a query coordinate \(\mathbf{x}\) and the contextualized latent pointcloud~(see Appendix~\ref{app:cross_attention}). Since \(\mathbf{x}\) has no content feature, this operation retains the positional query of ENF. We use the same nearest-neighbour and Gaussian windowing mechanisms and add a residual GeGLU block~\citep{shazeer2020glu} before the final output projection as per~\cite{zhang20233dshape2vecset}. Since all pose-dependent interactions are functions of bi-invariant attributes, the self-attention and cross-attention operations preserve equivariance under \(G\) (see Appendix~\ref{app:equivariance}). 

Self-attention changes both how the adapted latents are decoded and the gradient field through which they are inferred. Since each prediction depends on interactions among latent variables, an inner-loop update to one latent can incorporate information represented by its neighbors. MAML differentiates through these coupled updates, allowing the outer objective to shape how local latents should coordinate to form a coherent representation within a small number of adaptation steps.
\section{Experiments}
\label{sec:experiments}

We conduct two sets of experiments to evaluate MetaLF. First, we use polynomial fields with known underlying functions to study how self-attention shapes meta-learned latent representations and helps capture structure beyond local observations. We subsequently evaluate MetaLF on image reconstruction, SDF fitting, meta-classification, and meta-segmentation, testing reconstruction quality and end-to-end learning of downstream objectives. Across experiments, we compare against Functa~\citep{dupont2022data}, ENF~\citep{wessels2025grounding}, and additional relevant baselines.

\subsection{Polynomial fields}
For a behavioral analysis of MAML we consider the task of polynomial field reconstruction up to degree six, for which the function family is described by
\[
f_c(x,y)=\sum_{p+q\leq 6} c_{pq}x^py^q,
\qquad (x,y)\in[-1,1]^2.
\]
Defining a known function family allows us to investigate both reconstruction quality and the geometry of the learned optimization encoder. For this formal analysis, we use a latent set of $N=25$ and evaluate how latent dynamics change with self-attention depth. We report reconstruction mean squared error~(MSE), the effective rank $r_{\mathrm{eff}}$ of the latent spectrum~\citep{roy2007effective}, and the polynomial tangent fraction \(\tau_{\mathrm{poly}}\), which measures the alignment of local decoder variations with the ground-truth polynomial function space. Formal definitions and further implementation details are provided in Appendix~\ref{app:polynomial_implementation}. We additionally compare FOMAML and MAML training dynamics in Appendix~\ref{app:maml_vs_fomaml}.

\begin{wraptable}[12]{r}
{0.65\textwidth}
\centering
\vspace{-0.7cm}
\caption{MetaLF self-attention improves both encoding structure and reconstruction quality. Increasing $N_{\mathrm{SA}}$ yields lower-dimensional, more polynomial-aligned encodings (effective rank $r_{\mathrm{eff}}$, tangent fraction $\tau_{\mathrm{poly}}$) and higher decoded-field PSNR. Results are averaged over three seeds.}
\label{tab:poly_results}
\resizebox{0.65\columnwidth}{!}{
\begin{tabular}{lcccc}
\toprule
& \multicolumn{2}{c}{Full-grid PSNR $\uparrow$ (dB)}
& \multicolumn{2}{c}{MAML geometry} \\
\cmidrule(lr){2-3}\cmidrule(lr){4-5}
 & MAML & FOMAML
& $r_\mathrm{eff}$ $\downarrow$
& $\tau_{\mathrm{poly}}$ $\uparrow$ \\
\midrule
Functa       & 37.31 (0.31) & 35.03 (1.32) & 128.27 (2.11)  & 0.539 (0.006) \\
ENF          & 44.15 (0.56) & 32.95 (0.65) & 118.14 (13.54) & 0.418 (0.031) \\
\midrule
\rowcolor{gray!30}
MetaLF ($N_{\mathrm{SA}}=0$) & 42.28 (0.13) & 33.25 (0.62) & 72.85 (4.78)   & 0.619 (0.035) \\
\rowcolor{gray!30}
MetaLF ($N_{\mathrm{SA}}=2$) & 63.79 (1.35) & 37.40 (1.87) & 33.31 (0.29)   & 0.837 (0.004) \\
\rowcolor{gray!30}
MetaLF ($N_{\mathrm{SA}}=4$) & \textbf{64.24} (0.82) & \textbf{38.30} (0.76) & \textbf{31.09} (0.31)   & \textbf{0.861} (0.004) \\
\bottomrule
\end{tabular}}
\end{wraptable}

Table~\ref{tab:poly_results} reports polynomial reconstruction quality and the corresponding latent-space geometry under MAML optimization. As self-attention depth increases, the effective rank of MetaLF's latent representation seems to converge toward 28, which is the minimum dimensionality required to describe the degree-six polynomial family, indicating that self-attention reorganizes the latent space toward a compact basis aligned with the underlying function class. This geometric compression coincides with a monotonic improvement in PSNR, with MetaLF ($N_{\mathrm{SA}} = 4$) achieving the highest fidelity among all methods. Appendix Figure~\ref{fig:attention_pattern} further illustrates this effect, showing attention coupling extending across increasingly non-local latent neighborhoods as $N_{{SA}}$ grows.

\begin{wraptable}[10]{r}{0.47\textwidth}
\centering
\vspace{-0.6cm}
\caption{Test-set reconstruction PSNR~(dB,~$\uparrow$) for CIFAR10, CelebA (64x64) and ImageNet-1K (128x128), averaged over three seeds.}
\label{tab:image_recon}
\resizebox{0.47\columnwidth}{!}{
\begin{tabular}{llccc}
\toprule
 & & CIFAR10 & CelebA & ImageNet \\
\midrule
Functa                     & & 36.2 (0.1) & 32.6 (0.1) & 23.1 (0.0) \\
ENF & $a^{\mathbb{R}^2}$     & 42.5 (0.3) & 35.7 (0.1) & 27.7 (0.0) \\
ENF & $a^{\mathrm{SE}(2)}$   & 42.4 (0.2) & 35.7 (0.1) & 27.6 (0.0) \\
\midrule
\rowcolor{gray!30}
MetaLF & $a^{\mathbb{R}^2}$   & 47.7 (0.2) & 39.4 (0.0) & \textbf{32.0} (0.0) \\
\rowcolor{gray!30}
MetaLF & $a^{\mathrm{SE}(2)}$ & \textbf{47.9} (0.1) & \textbf{39.5} (0.1) & 31.9 (0.0) \\
\bottomrule
\end{tabular}}
\end{wraptable}

\subsection{End-to-end meta-learning}
Here we apply MetaLF to a number of meta-learning tasks, evaluating its ability to construct useful representations through an expressive optimization encoder. We consider signal reconstruction and SDF fitting as inner-outer objective pairs that share the same reconstruction loss, as well as meta-classification and meta-segmentation, where the outer objective diverges from the inner reconstruction target and instead supervises a downstream semantic task in end-to-end fashion. We detail the per-task training setup in Appendix~\ref{app:training_setup}.

\paragraph{Image reconstruction} For image reconstruction, we consider the CIFAR10~\citep{krizhevsky2009learning}, CelebA~\citep{liu2015deep}, and ImageNet-1K~\citep{deng2009imagenet} datasets. We set $N_{\mathrm{SA}} = 4$ and compare reconstruction quality at matched latent budgets with $K = 3$ inner steps. As shown in~ Table~\ref{tab:image_recon}, MetaLF outperforms both Functa and ENF under both bi-invariant settings, achieving near-lossless reconstruction on CIFAR10 and substantial gains on the higher-resolution CelebA and ImageNet benchmarks. Notably, MetaLF shows little sensitivity to the choice of geometric group.

The small differences between group parameterizations also appear in ENF, while MetaLF consistently improves reconstruction under either choice. We additionally provide an ablation of the MetaLF parameterization and an analysis of computational costs in Appendix~\ref{app:latent_ablation}, and evaluate MetaLF in a generative setting in Appendix~\ref{app:generative}.
% This suggests that self-attention compensates for the reduced equivariance constraints of the translation-only group, allowing the latent point set to recover much of the benefit that rotation-awareness otherwise provides in ENF. An ablation on ATLAS parameterization and an analysis on computation costs are presented in Appendix~\ref{app:latent_ablation}. We further test ATLAS in a generative setting in Appendix~\ref{app:generative}.

\begin{wraptable}[11]{r}{0.65\textwidth}
\centering
\vspace{-0.1cm}
\caption{MAML-based shape reconstruction IoU for the ShapeNet-Part, -Core, and ACDC hold-out test-sets. Values are the mean (standard deviation) over three seeds.}
\label{tab:shape_recon}
\resizebox{0.65\columnwidth}{!}{
\begin{tabular}{lccccc}
\toprule
& \multicolumn{2}{c}{ShapeNet IoU $\uparrow$}
& \multicolumn{3}{c}{ACDC IoU $\uparrow$} \\
\cmidrule(lr){2-3}\cmidrule(lr){4-6}
 & Part & Core & LV & Myo & RV \\
\midrule
Functa & 4.0 (0.1) & 5.0 (0.0) & 1.2 (0.0) & 1.1 (0.0) & 1.2 (0.0) \\
SpatialFuncta & 49.0 (2.2) & 46.2 (2.3) & 69.4 (5.5) & 47.2 (2.6) & 55.5 (2.0) \\
ENF & 65.5 (0.5) & 67.2 (1.9) & 77.1 (2.1) & 37.3 (2.6) & 64.7 (1.1) \\
\rowcolor{gray!30}
MetaLF & \textbf{76.7} (1.5) & \textbf{78.2} (4.0) & \textbf{89.3} (0.5) & \textbf{77.9} (0.4) & \textbf{79.5} (2.7) \\
\bottomrule
\end{tabular}}
\end{wraptable}

\paragraph{Shape reconstruction} For shape reconstruction, we consider 3D models from the ShapeNet-Part~\citep{yi2016scalable} and ShapeNet-Core~\citep{yu2021pointr} subsets of ShapeNet~\citep{chang2015shapenet}. We additionally consider cardiac shapes derived from 3D MRI segmentations of the left ventricle (LV) blood pool, LV myocardium (Myo), and the right ventricle (RV)~(ACDC~\citep{bernard2018deep}), and describe any preprocessing in Appendix~\ref{app:shape_recon}. For all tasks, reconstruction is optimized by fitting per-shape SDFs using $K=5$ inner steps, with all three cardiac shapes being rendered simultaneously from the same adapted latent representation. Table~\ref{tab:shape_recon} shows that MetaLF ($N_{\mathrm{SA}} = 2$) consistently improves shape reconstruction, indicating that aggregating information beyond local latent neighborhoods helps capture globally coherent geometry while preserving local detail. The large gain for the thin, connected myocardium highlights the value of non-local coordination.
% Appendix: note that ENF ShapeNet-Core reaches 73.2 IoU under similar latent capacity settings, noticably lower still

\begin{wraptable}[12]{r}{0.65\textwidth}
\centering
\vspace{-0.5cm}
\caption{MAML-based classification and reconstruction on the CIFAR10, ImageNet-1K, and ShapeNet-Core hold-out test sets. PSNR~(dB) and IoU are mean (standard deviation) over test items; image accuracy is top-1 (binomial standard error) and ShapeNet accuracy is macro-averaged over its 55 classes.}
\label{tab:meta_cls}
\resizebox{0.65\columnwidth}{!}{
\begin{tabular}{lcccccc}
\toprule
& \multicolumn{2}{c}{ImageNet}
& \multicolumn{2}{c}{CIFAR10}
& \multicolumn{2}{c}{ShapeNet-Core} \\
\cmidrule(lr){2-3}\cmidrule(lr){4-5}\cmidrule(lr){6-7}
 & PSNR $\uparrow$ & Acc. $\uparrow$ & PSNR $\uparrow$ & Acc. $\uparrow$ & IoU $\uparrow$ & Acc. $\uparrow$ \\
\midrule
MWT & 21.8 (-) & 24.1 (-) & 30.9 (-) & 64.7 (-) & - & - \\
\midrule
Functa        & 21.6 (2.5) & 7.4 (0.1) & 23.1 (1.9) & 57.7 (0.5) & 25.6 (8.0) & 1.8 (9.4) \\
SpatialFuncta & 25.5 (2.5) & 4.1 (0.1) & 36.5 (2.9) & 42.6 (0.5) & 48.6 (12.9) & 64.8 (30.8) \\
ENF           & 26.3 (2.7) & 4.5 (0.1) & \textbf{39.2} (3.3) & 34.4 (0.5) & 48.8 (13.2) & 57.3 (34.3) \\
\rowcolor{gray!30}
MetaLF         & \textbf{27.9} (2.9) & \textbf{42.6} (0.2) & 37.8 (2.4) & \textbf{83.3} (0.4) & \textbf{55.3} (16.9) & \textbf{70.8} (24.7) \\
\bottomrule
\end{tabular}}
\end{wraptable}

\paragraph{Meta-classification} In meta-classification, we train an additional linear classifier on top of the pre-CA intermediate latent representation. A cross-entropy classification loss is backpropagated from this head together with the standard reconstruction loss during the outer step only, i.e. \mbox{\(\mathcal{L}_{\mathrm{out}} = \mathcal{L}_{\mathrm{rec}} + \mathcal{L}_{\mathrm{CE}}\)}. We test this end-to-end meta-learning setting for CIFAR10, ImageNet-1K and ShapeNet-Core classification, and list results in Table~\ref{tab:meta_cls}. While classification gains on ShapeNet-Core are modest, MetaLF achieves substantially higher accuracy on CIFAR10 (83.3\%), and performance separates further with scale: MetaLF ($N_{\mathrm{SA}} = 4$) reaches 42.6\% top-1 accuracy on ImageNet-1K against 4--7\% for baselines. MetaLF also substantially outperforms the strongest prior end-to-end meta-classification work on both image datasets (MWT~\citep{gielisse2025end}). We provide a more extensive discussion on these results in Appendix~\ref{app:cls_discussion}.

\begin{wraptable}[19]{r}{0.5\textwidth}
\centering
\vspace{-0.6cm}
\caption{Meta-learned segmentation on the OMBRIA, ShapeNet-Part,
and OASIS hold-out test sets. Header $K$ denotes test-time fitting
steps for our methods. Scores are the pooled flood IoU
for OMBRIA, instance mIoU for ShapeNet-Part, and five-class
foreground Dice for OASIS. Parentheses denote standard
deviations over test items. Failed denotes non-fit/divergence under the attempted recipes.}
\label{tab:meta_seg}
\resizebox{\linewidth}{!}{
\begin{tabular}{lccc}
\toprule
& OMBRIA & ShapeNet-Part & OASIS \\
& ($K=3$) & ($K=5$) & ($K=5$) \\
\cmidrule(lr){2-2}\cmidrule(lr){3-3}\cmidrule(lr){4-4}
& IoU $\uparrow$ & mIoU $\uparrow$ & Dice $\uparrow$ \\
\midrule
OmbriaNet & 72.4 & -- & -- \\
MetaSeg ($K=100$) & -- & -- & 91.0 (1.1) \\
NISF & -- & -- & 81.0 (0.7) \\
\midrule
Functa & 39.9 & Failed & Failed \\
SpatialFuncta & 68.9 & 78.7 (19.1) & 78.2 (1.1) \\
ENF & 69.9 & 81.2 (17.5) & 83.9 (1.1) \\
\rowcolor{gray!30}
MetaLF & \textbf{72.3} & \textbf{83.4} (16.7)
      & \textbf{86.9} (0.9) \\
\bottomrule
\end{tabular}}
\end{wraptable}

\paragraph{Meta-segmentation} We extend the decoder with segmentation outputs and jointly optimize reconstruction and segmentation in the outer loop, while retaining reconstruction-only adaptation at test time. We evaluate the same method across diverse segmentation tasks on OMBRIA~\citep{drakonakis2022ombrianet}, ShapeNet-Part~\citep{yi2016scalable}, and OASIS~\citep{marcus2007open}, covering 2D images, 3D shapes, and 3D images, respectively, with \(N_{\mathrm{SA}} = (4, 2, 4)\) and \mbox{\(K = (3, 5, 5)\)}. Table 5 shows that MetaLF consistently outperforms the evaluated few-step baselines across all three datasets. On OASIS, MetaLF achieves 86.9\% foreground Dice, exceeding the reported performance of the implicit segmentation method NISF~\citep{stolt2023nisf} (81.0\%) and approaching MetaSeg~\citep{vyas2025fit} (91.0\%) with only five fitting steps, compared with 100 for MetaSeg. Although these published results use different protocols, they place MetaLF within reach of specialized segmentation methods at a small adaptation budget. These results demonstrate that MetaLF can learn an effective semantic representation through reconstruction updates alone, extending the benefit of the optimization encoder perspective beyond signal fitting to dense prediction. Details on meta-segmentation are described in Appendix~\ref{app:meta_seg}.

\paragraph{Limitations and future work}
MetaLF benefits most from datasets with shared underlying structure as exemplified in Table~\ref{tab:shape_recon} (ACDC) and Appendix~\ref{app:generative} (CelebA), while performance gains are more modest for heterogenous data. This suggests that latent attention is most effective when the decoder can learn common structure across samples. Equivariant self-attention further introduces a computational tradeoff, where increasing $K_{\mathrm{SA}}$ requires additional attention matrices to account for each neighbour's relative geometry. Although broader interactions allow for more informed updates, we found that global attention resulted in overly smooth reconstructions, while localized attention improved reconstruction quality and training stability. Exploring more flexible attention neighbourhoods may help balance these effects. Beyond attention, second-order training and iterative test-time latent optimization remain computational costs even when only a few updates are required. Future work should further examine how decoder design, inner- and outer-loop field sampling, and latent pointcloud size affect this balance. MetaLF represents one design choice informed by the optimization encoder perspective, providing a basis for investigating these choices.
\section{Conclusion}

In this work, we formalize few-step adaptation of latent variables as an optimization encoder, showing how second-order meta-learning trains the process that constructs a representation alongside the decoder that uses it. This perspective makes decoder architecture a design choice for inference itself and directly motivates MetaLF: equivariant self-attention coordinates spatially grounded latents, shaping the gradient updates through which local observations inform a coherent representation. Our controlled analysis links this coordination to more compact, function-aligned latent geometry, while our experiments demonstrate improved few-step image and shape reconstruction. By separating the inner encoding objective from outer task supervision, MetaLF also learns representations for classification and segmentation in an end-to-end fashion through reconstruction-only adaptation at test time. Together, these results establish optimization encoding as a practical basis for designing neural fields that capture local detail, coordinate non-local structure, and support semantic prediction within a small number of gradient updates.

\section*{Acknowledgments}
This study was supported by ZonMw Rubicon under grant no.~04520252520006.

\bibliographystyle{unsrt}
\bibliography{references}

\clearpage
\appendix
\section{Appendix}

\subsection{Meta-learning training loop}
\label{app:meta_learning}
In a standard neural field setting without an instance-specific latent code, the inner loop adapts a copy of the neural-field parameters \(\theta\) to each sample, while the outer loop updates their shared initialization. MAML differentiates through the full sequence of inner-loop updates, whereas FOMAML computes the gradient at the final adapted parameters while ignoring the optimization trajectory.

Conditional neural fields instead associate each sample \(i\) with a latent representation \(z_i\). Following CAVIA~\citep{zintgraf2019fast}, the inner loop adapts only \(z_i\), starting from a shared initialization \(z^0\), while the outer loop updates the shared decoder parameters \(\theta\). Although \(\theta\) remains fixed during inner-loop adaptation, it determines the gradients used to update \(z_i\). As shown in Algorithm~\ref{alg:meta_learning_cnf}, MAML retains this dependence and differentiates through the latent updates, while FOMAML stops the gradient through the adaptation trajectory.

\begin{algorithm}[h]
\caption{Meta-learning a conditional neural field}
\label{alg:meta_learning_cnf}
\begin{algorithmic}[1]
\Require Inner learning rate $\alpha$, outer learning rate $\eta$,
number of inner steps $K$, method
$m\in\{\textsc{MAML},\textsc{FOMAML}\}$
\State Initialize the shared decoder $f_\theta$
and latent initialization
$z^0=\{(p_n,c_n)\}_{n=1}^{N}$
\While{not converged}
    \State Sample a batch of signals $\{f_i\}_{i\in\mathcal B}$
    \ForAll{$i\in\mathcal B$}
        \State Sample inner and outer observations
        $\mathcal D_i^{\mathrm{in}}$ and $\mathcal D_i^{\mathrm{out}}$
        \State Initialize $z_i^0 \gets z^0$
    \EndFor
    \For{$k=0,\ldots,{K}-1$}
        \ForAll{$i\in\mathcal B$}
            \State $\widetilde z_i^{k+1}
            \gets z_i^k-\alpha\nabla_{z_i^k}
            \mathcal L_{\mathrm{in}}
            (\theta,z_i^k;\mathcal D_i^{\mathrm{in}})$
            \If{$m=\textsc{FOMAML}$}
                \State $z_i^{k+1}\gets
                \operatorname{sg}[\widetilde z_i^{k+1}]$
                \Comment{stop gradient through adaptation}
            \Else
                \State $z_i^{k+1}\gets\widetilde z_i^{k+1}$
                \Comment{retain the adaptation graph}
            \EndIf
        \EndFor
    \EndFor
    \State $\displaystyle
    \mathcal L_{\mathrm{meta}}
    \gets \frac{1}{|\mathcal B|}
    \sum_{i\in\mathcal B}
    \mathcal L_{\mathrm{out}}
    (\theta,z_i^{K};
    \mathcal D_i^{\mathrm{out}})$
    \State $\theta\gets\theta-\eta\nabla_\theta
    \mathcal L_{\mathrm{meta}}$
\EndWhile
\end{algorithmic}
\end{algorithm}

\subsection{Differentiating through the optimization encoder}
\label{app:expanded_meta_gradient}

Here we derive the gradient through the complete latent adaptation that describes MAML. For
sample $i$, recall the inner update
\begin{equation}
    z_i^{k+1}
    = z_i^k
    - \alpha \nabla_z
    \mathcal{L}_{\mathrm{in}}^i(\theta,z_i^k),
    \qquad k=0,\ldots,K-1,
    \label{eq:appendix-inner-update}
\end{equation}
where dependence on the inner observations is omitted for readability. The
resulting optimization encoder is the composition
$E_\theta^K = U_{\theta,\mathcal D_i}^{\circ K}$ which returns
$z_i^K=E_\theta^K(\mathcal D_i)$. Although $\theta$ is not updated in the
inner loop, every update in Eq.~\ref{eq:appendix-inner-update} depends on it
through the gradient of the inner objective.

Let
\begin{equation}
    P_i^k := \frac{\partial z_i^k}{\partial \theta},
    \qquad
    H_i^k := \nabla_{zz}^2
    \mathcal{L}_{\mathrm{in}}^i(\theta,z_i^k),
    \qquad
    B_i^k := \nabla_{z\theta}^2
    \mathcal{L}_{\mathrm{in}}^i(\theta,z_i^k).
    \label{eq:appendix-sensitivity-definitions}
\end{equation}
Applying the chain rule to one inner update gives
\begin{align}
    P_i^{k+1}
    &= P_i^k
    - \alpha \frac{\partial}{\partial\theta}
      \left[
      \nabla_z\mathcal{L}_{\mathrm{in}}^i
      (\theta,z_i^k)
      \right] \nonumber\\
    &= P_i^k
    - \alpha\left(B_i^k + H_i^k P_i^k\right) \nonumber\\
    &= A_i^k P_i^k - \alpha B_i^k,
    \qquad A_i^k := I-\alpha H_i^k.
    \label{eq:appendix-sensitivity-recurrence}
\end{align}
The mixed derivative $B_i^k$ introduces the dependence on the decoder parameters
at step $k$, while $A_i^k$ transports all dependencies accumulated at earlier
steps through the next update. Repeated substitution in Eq.~\ref{eq:appendix-sensitivity-recurrence} yields
\begin{equation}
    P_i^K
    =
    A_i^{K-1}A_i^{K-2}\cdots A_i^0 P_i^0
    - \alpha
    \sum_{t=0}^{K-1}
    A_i^{K-1}A_i^{K-2}\cdots A_i^{t+1}B_i^t,
    \label{eq:appendix-unrolled-sensitivity}
\end{equation}
where the product preceding $B_i^{K-1}$ is the identity. When the shared initialization
$z^0$ is independent of $\theta$, $P_i^0=0$, so
\begin{equation}
    P_i^K
    = -\alpha
    \sum_{t=0}^{K-1}
    A_i^{K-1}A_i^{K-2}\cdots A_i^{t+1}B_i^t.
    \label{eq:appendix-unrolled-sensitivity-independent-init}
\end{equation}

Now let
$\mathcal{L}_{\mathrm{out}}^i(\theta,z_i^K)$ denote the outer objective. Its
total derivative with respect to the decoder parameters is
\begin{equation}
    \nabla_\theta \mathcal{L}_{\mathrm{out}}^i
    =
    \underbrace{
    \partial_\theta \mathcal{L}_{\mathrm{out}}^i
    }_{\text{direct decoder gradient}}
    +
    \underbrace{
    (P_i^K)^\top
    \nabla_z \mathcal{L}_{\mathrm{out}}^i
    }_{\text{gradient through the optimization encoder}}.
    \label{eq:appendix-meta-gradient-decomposition}
\end{equation}
Substituting Eq.~\ref{eq:appendix-unrolled-sensitivity-independent-init}
into Eq.~\ref{eq:appendix-meta-gradient-decomposition} gives the expanded
meta-gradient
\begin{align}
    \nabla_\theta \mathcal{L}_{\mathrm{out}}^i
    ={}& \partial_\theta \mathcal{L}_{\mathrm{out}}^i \nonumber\\
    &-\alpha\sum_{t=0}^{K-1}
    (B_i^t)^\top
    (A_i^{t+1})^\top\cdots(A_i^{K-1})^\top
    \nabla_z \mathcal{L}_{\mathrm{out}}^i.
    \label{eq:appendix-expanded-meta-gradient}
\end{align}
Thus, each term represents a dependency introduced through \(B_i^t\) at inner step \(t\) and propagated through every subsequent update before reaching the outer objective. Equivalently, the same derivative can be written in reverse mode. Define the
adjoint at the final representation as
\begin{equation}
    q_i^K := \nabla_z
    \mathcal{L}_{\mathrm{out}}^i(\theta,z_i^K),
\end{equation}
and propagate it backwards through the inner updates according to
\begin{equation}
    q_i^k = (A_i^k)^\top q_i^{k+1},
    \qquad k=K-1,\ldots,0.
    \label{eq:appendix-adjoint-recurrence}
\end{equation}
The decoder gradient is then
\begin{equation}
    \nabla_\theta \mathcal{L}_{\mathrm{out}}^i
    = \partial_\theta \mathcal{L}_{\mathrm{out}}^i
    - \alpha\sum_{k=0}^{K-1}(B_i^k)^\top q_i^{k+1},
    \label{eq:appendix-reverse-meta-gradient}
\end{equation}
which is identical to Eq.~\ref{eq:appendix-expanded-meta-gradient}. If the
shared initialization $z^0$ is learned, its gradient follows from the same
backward recursion:
\begin{equation}
    \nabla_{z^0}\mathcal{L}_{\mathrm{out}}^i
    = q_i^0
    = (A_i^0)^\top\cdots(A_i^{K-1})^\top
      \nabla_z\mathcal{L}_{\mathrm{out}}^i.
    \label{eq:appendix-initialization-gradient}
\end{equation}

Full MAML retains both components of the inner-update Jacobian: the
Hessian-dependent transport $A_i^k$ and the mixed-derivative injection
$B_i^k$. In the first-order approximation, the adaptation trajectory
is detached when computing the outer gradient. Equivalently for the decoder
gradient, this makes $A_i^k\approx I$ and $B_i^k\approx0$. Since $P_i^0=0$ with
respect to $\theta$, this gives $P_i^K\approx0$, and the second term in
Eq.~\ref{eq:appendix-meta-gradient-decomposition} vanishes. The decoder
therefore receives only
$\partial_\theta\mathcal{L}_{\mathrm{out}}^i$: first-order training executes
the same forward optimization encoder but does not backpropagate the outer
objective through the computation that constructed its representation.

\subsection{Extended description of MetaLF}

\subsubsection{Cross-attention}
\label{app:cross_attention}
While the main contributions of MetaLF lie in its design from an optimization encoder perspective, its mapping from latent features to field values builds on the ENF decoder of~\citep{wessels2025grounding}. Equivariant cross-attention with field coordinates $\mathbf{x}$ largely follows their implementation:
\[
    q_{j\to\mathbf{x}} = W_q\phi_q(a(\mathbf{x},p_j)),
    \qquad
    k_j = W_k\bar h_j^L,
    \qquad
    \bar h_j^L = \mathrm{LN}(h_j^L).
\]
Here, $h_j^L$ is the contextualized feature of latent $j$ after the self-attention stack. Since a field coordinate has no content feature, its query is constructed from its position relative to each latent pose. The relative attribute is $a(\mathbf{x},p_j)=\mathbf{x}-p_j$ for translations and $a(\mathbf{x},p_j)=R_j^\top(\mathbf{x}-t_j)$ for roto-translations.

Values use the same geometry-conditioned FiLM parameterization as in self-attention, now evaluated between field coordinates and latent poses. We also retain the Gaussian spatial window and normalize attention over the $K_{\mathrm{CA}}$ nearest latents, denoted by $\mathcal N_{\mathrm{CA}}(\mathbf{x})$. A single cross-attention head aggregates their contributions:
\[
    u(\mathbf{x})
    =
    W_o\sum_{j\in\mathcal N_{\mathrm{CA}}(\mathbf{x})}
    \alpha_{j\to\mathbf{x}}\,v_{j\to\mathbf{x}} + b_o,
\]
where $\alpha_{j\to\mathbf{x}}$ denotes the normalized attention weight. To increase the expressivity of the decoder, we further apply a residual GeGLU feed-forward block to the aggregated feature before the final output projection as done by~\citep{zhang20233dshape2vecset}:
\[
    \tilde u(\mathbf{x})
    =
    u(\mathbf{x})
    +
    \mathrm{FFN}_{\mathrm{CA}}\bigl(\mathrm{LN}(u(\mathbf{x}))\bigr),
    \qquad
    f_\theta(\mathbf{x};z)
    =
    W_{\mathrm{out}}\tilde u(\mathbf{x}) + b_{\mathrm{out}}.
\]
This block follows the same structure as the feed-forward blocks in the self-attention stack. It adds a pointwise nonlinear refinement after spatial aggregation while preserving equivariance.

\begin{figure*}[t]
\centering
     \includegraphics[width=\textwidth, trim=0 0.1cm 0 0]{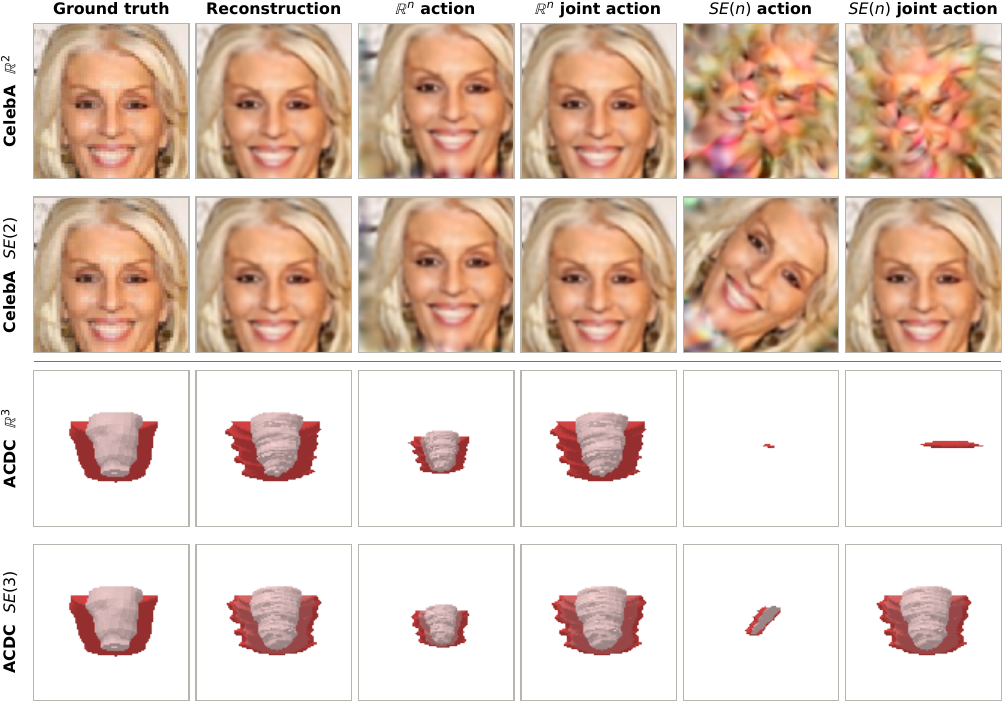}
     \caption{MetaLF is equivariant under a group action defined by its bi-invariant feature $a$. Rows show translation-equivariant (\(\mathbb{R}^n\)) and roto-translation-equivariant (\(\mathrm{SE}(n)\)) MetaLF models on CelebA and ACDC myocardium. Columns show ground truth, reconstruction in the canonical frame, pose-only translation, joint translation of poses and queries, pose-only roto-translation, and the corresponding joint roto-translation. Acting on poses alone transforms the represented field; applying the same group element jointly to poses and queries preserves predicted values at corresponding coordinates when the model respects that symmetry. Accordingly, joint translations match the canonical reconstruction in every row, whereas joint roto-translation does so only for the \(\mathrm{SE}(n)\) models.}
     \label{fig:equivariance}
\end{figure*}

\subsubsection{Equivariance}
\label{app:equivariance}
Equivariance of the decoded field follows from invariance under a joint transformation of the query coordinate and latent poses. For $z=\{(p_j,c_j)\}_{j=1}^N$, let $gz=\{(gp_j,c_j)\}_{j=1}^N$: the group acts on poses, while content channels remain unchanged. We show that
\[
    f_\theta(g\mathbf{x};gz)=f_\theta(\mathbf{x};z),
    \qquad\text{equivalently}\qquad
    f_\theta(\mathbf{x};gz)=f_\theta(g^{-1}\mathbf{x};z).
\]

The initial features $h_j^0=W_c c_j$ are unchanged by this action. Suppose that the features entering a self-attention block are also unchanged. Since
\[
    a(gp_j,gp_{j'})=a(p_j,p_{j'}),
\]
all queries, keys, and values computed from these features and attributes remain unchanged. Spatial windowing is likewise unchanged when expressed through these attributes, and Euclidean nearest-neighbour sets are preserved under translations and rotations. Attention therefore produces the same aggregated features. Layer normalization, GeGLU, and residual additions operate only on feature channels, so this property holds throughout the self-attention stack by induction.

For cross-attention, bi-invariance gives
\[
    a(g\mathbf{x},gp_j)=a(\mathbf{x},p_j).
\]
The positional queries, content-derived keys, and geometry-conditioned values are consequently unchanged, as are their attention weights. Thus $u(g\mathbf{x};gz)=u(\mathbf{x};z)$. The pointwise GeGLU block and final projection preserve this equality, establishing the claimed equivariance. Below we describe 2D and 3D equivariance in detail, and we demonstrate the behaviour in Figure~\ref{fig:equivariance} using examples from the CelebA~\citep{liu2015deep} and ACDC~\cite{bernard2018deep} datasets.

\paragraph{2D example}
For translations, $p_j\in\mathbb{R}^2$ and
$a(\mathbf{x},p_j)=\mathbf{x}-p_j$. Translating both coordinates and poses by $\mathbf{b}$ leaves this difference unchanged:
\[
    (\mathbf{x}+\mathbf{b})-(p_j+\mathbf{b})
    =\mathbf{x}-p_j.
\]
For roto-translations, let $p_j=(t_j,R_j)$ with $R_j\in\mathrm{SO}(2)$, and use
$a(\mathbf{x},p_j)=R_j^\top(\mathbf{x}-t_j)$.
Under $g=(Q,\mathbf{b})\in\mathrm{SE}(2)$,
\[
    \mathbf{x}'=Q\mathbf{x}+\mathbf{b},
    \qquad
    t_j'=Qt_j+\mathbf{b},
    \qquad
    R_j'=QR_j.
\]
Then
\[
    (R_j')^\top(\mathbf{x}'-t_j')
    =R_j^\top Q^\top Q(\mathbf{x}-t_j)
    =R_j^\top(\mathbf{x}-t_j).
\]
The self-attention attributes
$R_{j'}^\top(t_j-t_{j'})$ and $R_{j'}^\top R_j$
are unchanged by the same calculation. Transforming the latent poses therefore translates or rotates the decoded image while preserving its channel values.

\paragraph{3D example}
Translation equivariance in $\mathbb{R}^3$ follows from the same cancellation. For roto-translations, poses carry a three-dimensional frame $R_j\in\mathrm{SO}(3)$. Given $g=(Q,\mathbf{b})\in\mathrm{SE}(3)$, the cross-attention attribute satisfies
\[
    (QR_j)^\top
    \bigl(Q\mathbf{x}+\mathbf{b}-(Qt_j+\mathbf{b})\bigr)
    =R_j^\top(\mathbf{x}-t_j),
\]
and relative orientations satisfy
\[
    (QR_{j'})^\top(QR_j)=R_{j'}^\top R_j.
\]
Together with invariance of the relative translations, these identities establish bi-invariance for both attention operations. For a decoded signed distance field, rigidly transforming the latent poses therefore moves the represented shape: the field value at $g\mathbf{x}$ after transformation equals its value at $\mathbf{x}$ before transformation.

\subsubsection{General implementation details}
As observed by~\citep{wessels2025grounding}, we find that initialization of the latent poses $z=\{(p_j,c_j)\}_{j=1}^N$ typically benefits from maximum coverage of the output field. We therefore similarly initialize latents to a grid formation and break symmetry by adding a small amount of Gaussian noise~($\mathcal{N}(0, 1e-3)$). Given this configuration, we typically set $K_{\mathrm{CA}}$ to 4 and 8 for 2D and 3D fields respectively, thus defining the nearest neighbours of a query coordinate $\mathbf{x}$ during cross-attention. We further use
\(
    K_{\mathrm{SA}}=\sqrt[\mathrm{dim}]{\mathrm{N}},
\)
corresponding to \(\sqrt{N}\) neighbours in 2D and \(\sqrt[3]{N}\) in 3D. Optimization encoding of $z$ is learned together with meta-SGD~\citep{li2017meta}, which significantly boosts performance across experiments. Coordinate positions of latent poses are further stabilized during training using a tanh function, thus ensuring that latent positions do not diverge from the grid defined between -1 and 1. Finally, to accommodate MAML, we implement MetaLF and baselines in \href{https://docs.jax.dev/en/latest/}{JAX}.

We further note that FiLM modulation in MetaLF self-attention may use either the bi-invariant geometric feature $a\rightarrow r$ or the content feature $c\rightarrow u$ as modulation base, as their combined value shares the same expressive ceiling. However, we noticed that optimization using the geometric feature $r$ as a basis resulted in much more stable convergence, which we hypothesize is due to the feature being more consistent over the course of training, as the relative orientation of the latent pointcloud does not fluctuate severely. The influence of the content feature $u$ is thus zero at the start of training, and may be utilized more as training progresses as deemed necessary to fulfill the modeling objective.

\subsection{Controlled analysis}
\label{app:controlled_analysis}

\subsubsection{Polynomial-field implementation and geometry metrics}
\label{app:polynomial_implementation}

\paragraph{Fields and latent representation.}
We sample polynomial fields on $\mathcal X=[-1,1]^2$ from the space of polynomials of total degree at most six, of which the output space contains $\binom{2+6}{6} = 28$ monomials. Fields are represented on a $32\times32$ grid, while raw coefficients are sampled independently from $\operatorname{Unif}[-1,1]$ and scaled by $0.85^{p+q}$, with an additional factor of $0.5$ for the constant term. Both ENF~\citep{wessels2025grounding} and MetaLF use $N=25$ tokens with latent states initialized on a $5\times5$ grid, while Functa~\citep{dupont2022data} instead uses a flat $512$-dimensional latent. Further Functa meta-learning settings are as described in~\citep{bauer2023spatial}, i.e. a FiLM-modulated SIREN network~\citep{sitzmann2020implicit} with $\omega_0=30$, and 15 layers of width 512.

\paragraph{Training and evaluation protocol.}
Models are trained for $500$ epochs with batch size $32$. Each inner loop returns $256$ grid observations sampled uniformly without replacement. We apply $K=3$ inner updates and differentiate the outer loss through all three updates using full second-order MAML. MetaLF uses meta-SGD~\citep{li2017meta} to optimize inner learning rates, initialized with $0.1$ for pose and $0.01$ for content, Adam~\citep{kingma2014adam} with outer learning rate $10^{-4}$, and global gradient-norm clipping at $1$. 
During evaluation, each diagnostic field is fitted from a deterministic $256$-coordinate mask, and all geometry quantities are evaluated on the complete $32\times32$ grid.

\paragraph{Content Jacobian and update-scaled geometry.}
For one fitted field, let
\(
f_\theta(z_3)\in\mathbb R^{M}
\)
denote the decoder output on the complete grid. We hold the adapted poses fixed and differentiate only with respect to the content part of $z_3$, i.e. $z^c$:
\[
J_c
=
\left.
\frac{\partial f}{\partial z^{\mathrm{c}}}
\right|_{z_3}
\in\mathbb R^{M\times d}.
\]
Let $D_c >0$ be the diagonal matrix of content step sizes for $z^c$. Given that we use meta-SGD~\citep{li2017meta}, its diagonal contains the learned positive-valued content update learning rates. We then define the meta-SGD-scaled Gauss-Newton matrix $\widetilde G_c$ as
\[
W_c=J_cD_c^{1/2},
\qquad
\widetilde G_c
=
W_c^\top W_c
=
D_c^{1/2}J_c^\top J_cD_c^{1/2}.
\]

\paragraph{Latent effective rank.}
Let $\lambda_1,\ldots,\lambda_r$ be the eigenvalues of $\widetilde G_c$ retained under the numerical threshold
\(
\lambda_i>10^{-7}\lambda_{\max}.
\)
After normalizing the retained spectrum with
\(
p_i=\lambda_i/(\sum_{j=1}^{r}\lambda_j),
\)
we define the latent effective rank by
\[
r_{\mathrm{eff}}
=
\exp\left(-\sum_{i=1}^{r}p_i\log p_i\right).
\]
This quantity uses entropy to measure the effective number of retained singular modes of the update-scaled Jacobian. When the spectrum is flat across \(r\) modes, \(r_{\mathrm{eff}}=r\). When one mode dominates, \(r_{\mathrm{eff}}\) is close to one. The implementation sets \(r_{\mathrm{eff}}=0\) when \(W_c\) has zero energy.

\paragraph{Polynomial tangent fraction.}
Let
\[
\Phi\in\mathbb R^{1024\times28},
\qquad
\Phi_{m,j}=x_m^{p_j}y_m^{q_j},
\]
be the matrix of monomials of total degree at most six, evaluated on the full grid. The orthogonal projector onto the resulting polynomial subspace is
\[
P_{\mathrm{poly}}
=
\Phi(\Phi^\top\Phi)^\dagger\Phi^\top.
\]
We measure how much of the local decoder variation lies in this subspace using
\[
\tau_{\mathrm{poly}}
=
\frac{\|P_{\mathrm{poly}}W_c\|_F^2}
{\|W_c\|_F^2}
=
\frac{\operatorname{tr}
\left(W_c^\top P_{\mathrm{poly}}W_c\right)}
{\operatorname{tr}\left(W_c^\top W_c\right)}.
\]
This ratio lies between zero and one and gives the fraction of local output-change energy within the $28$-dimensional polynomial subspace. Values near one indicate that small content perturbations mainly produce changes within the polynomial family; values near zero indicate that these changes mainly fall outside it. We compute the projection using the discrete Euclidean inner product on the $32\times32$ grid samples, thus summarizing local variation across content directions.

\subsubsection{FOMAML and MAML training behaviour}
\label{app:maml_vs_fomaml}

\begin{figure*}[h]
\centering
     \includegraphics[width=\textwidth, trim=0 0.4cm 0 0.4cm]{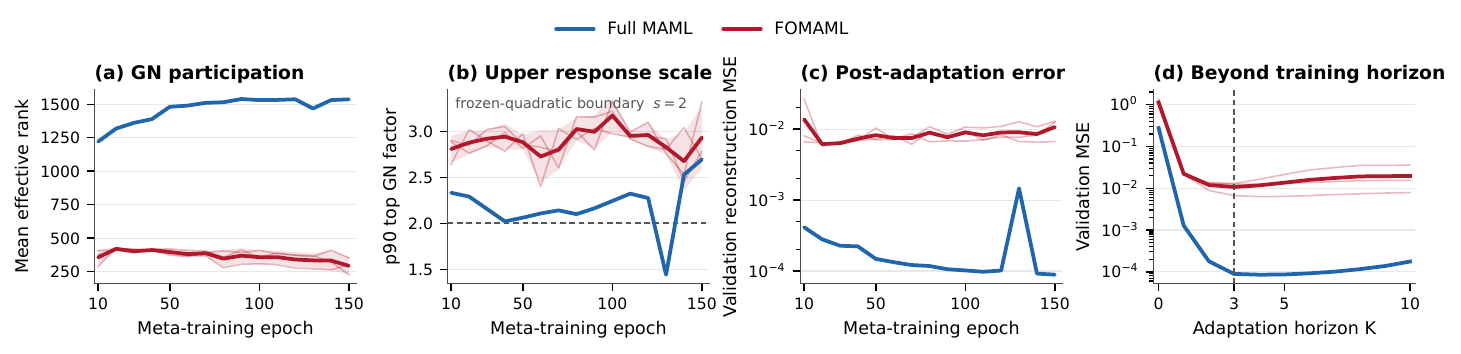}
     \caption{Training progress on CIFAR10 reconstruction for MAML (blue) and FOMAML (red). (a)~MAML develops a higher latent effective rank, while the first-order controls remain at lower ranks. (b) MAML generally has smaller upper response factors, although these fluctuate during training. (c) MAML's reconstruction error decreases substantially over training epochs, whereas the first-order controls plateau after an initial improvement. (d) Both methods perform best near the three-step training horizon, with reconstruction error gradually increasing under further adaptation.}
     \label{fig:maml_training}
\end{figure*}

We provide an analysis of the training dynamics on CIFAR10 reconstruction in Figure~\ref{fig:maml_training}. Compared to the simpler task of polynomial field reconstruction (see Table~\ref{tab:poly_results}), the latent effective rank gradually increases during MAML training on CIFAR10. This is consistent with the greater complexity and diversity of natural images, for which capturing variation may require responses spread across more latent directions. FOMAML instead maintains a much lower effective rank that gradually declines, alongside limited improvement in reconstruction quality. Stopping the gradient through the adapted latents prevents the outer objective from directly training the encoding procedure, which helps explain this difference. Full MAML retains this pathway, allowing reconstruction supervision to shape how the inner updates construct the latent representation.

The upper response scale measures how strongly an inner-loop update can respond along the most sensitive latent direction, accounting for both decoder sensitivity and the inner learning rate. The higher effective rank, together with a generally lower upper response scale, suggests that MAML spreads its local response more evenly across latent directions, with less concentration in the most sensitive directions. Finally, we measure how additional inner steps affect reconstruction beyond the three-step training horizon (Figure~\ref{fig:maml_training}d). MAML improves slightly with a fourth update, while FOMAML achieves its lowest mean error at three steps. Further updates gradually worsen reconstruction for both methods. This is consistent with the optimization encoder perspective: the decoder learns to use the structured latent representations produced within the training horizon, and additional updates may move these representations away from the distribution encountered during training.

\begin{figure*}[t]
\centering
     \includegraphics[width=\textwidth, trim=0 0.4cm 0 0]{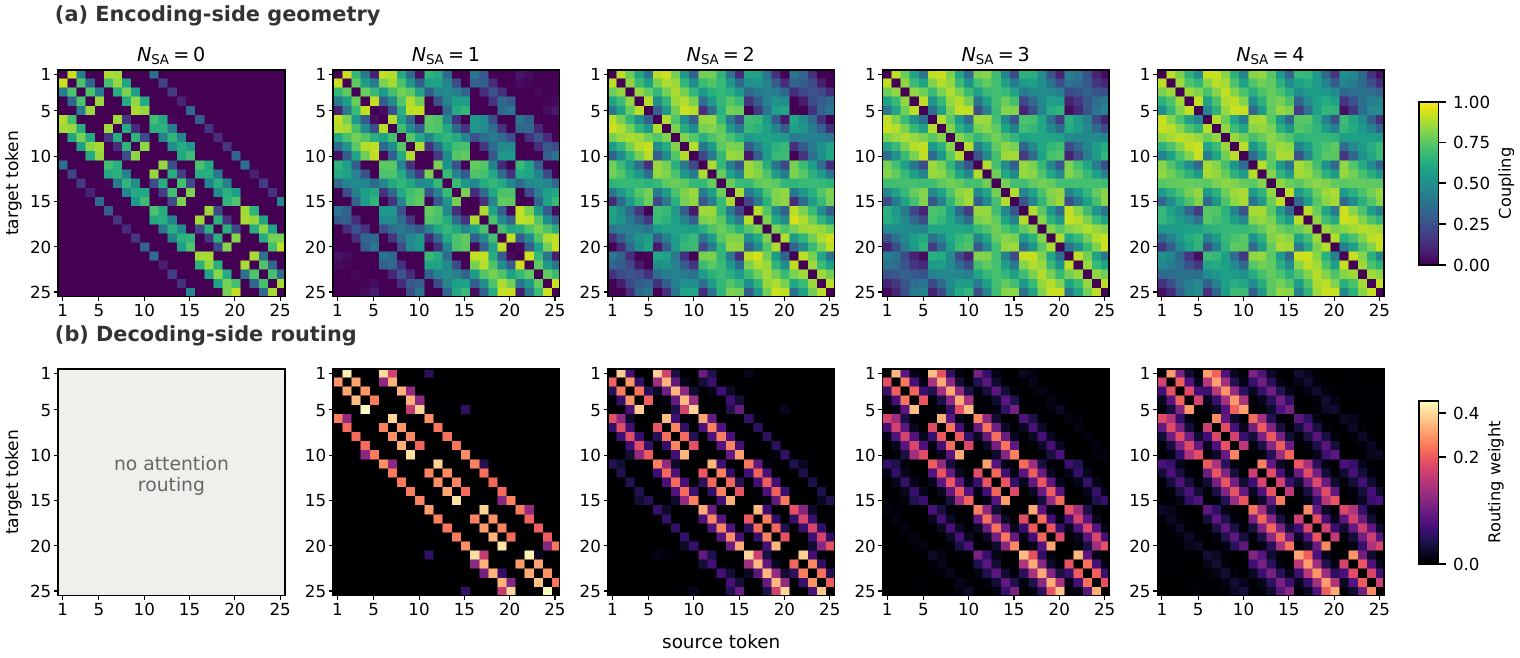}
     \caption{Optimization encoder-side coupling (top) and decoder-side attention routing (bottom) evaluated on the polynomial fields dataset, shown with increasing number of MetaLF self-attention layers ($N_{SA}$) for $N=25$ latent tokens. Additional layers allow information to propagate beyond local latent neighborhoods while retaining the spatial structure of the representation.}
     \label{fig:attention_pattern}
\end{figure*}

Figure~\ref{fig:attention_pattern} compares how tokens interact on the encoding and decoding sides as decoder depth increases. Without self-attention ($N_{\mathrm{SA}}=0$), each token affects only its local neighbourhood of the output field. Updates are therefore coupled only between spatially adjacent tokens, which results in the banded structure of the $5\times5$ token grid. Adding self-attention layers makes this coupling denser and less local: the mean off-diagonal coupling rises from $0.36$ at $N_{\mathrm{SA}}=0$ to $0.67$ at $N_{\mathrm{SA}}=4$. As a result, an inner-loop gradient step on one token increasingly interacts with updates on distant tokens. The decoding-side routing changes in a matching way. With a single attention layer, a token's non-self influence is routed almost entirely to its immediate neighbours. Deeper decoders spread this influence over progressively larger neighbourhoods: the off-diagonal routing mass grows from $0.26$ to $0.69$, and the normalized row entropy grows from $0.28$ to $0.63$ between $N_{\mathrm{SA}}=1$ and $N_{\mathrm{SA}}=4$. The two views therefore describe the same mechanism from opposite directions. As attention spreads information from each token further across the grid, the decoder's sensitivity to each latent also becomes less local, and the optimization encoder has to solve a more strongly coupled inner problem. Under this view, $N_{\mathrm{SA}}$ controls how far a local latent edit spreads through the reconstruction.

\subsection{Experimental details}
\label{app:training_setup}

\subsubsection{Image reconstruction}
\label{app:reconstruction_details}
For image reconstruction, we use MetaLF with $N_{\mathrm{SA}}=4$ self-attention layers, four attention heads, hidden width $128$, and $64$ content channels $c_j$ per latent token. We summarize the dataset-specific settings in Table~\ref{tab:MetaLF_image_training}, where $K_{\mathrm{SA}}$ and $K_{\mathrm{CA}}$ denote the number of nearest neighbours used for self-attention and cross-attention, respectively. The frequency scales control the positional embeddings used for query and value modulation.

\paragraph{Optimization}
Inner and outer objectives both minimize MSE. We use $K=3$ inner updates and differentiate through all three updates
using MAML. The shared latent initialization is learned together with the decoder, while meta-SGD learns a separate learning rate for
each latent coordinate and each inner step, initialized to $0.1$ for pose and $0.2$ for content. These rates are optimized using a separate Adam optimizer with learning rate $0.2$ and constrained to $[10^{-6},10]$. The decoder and shared initialization use Adam with the outer learning rates in Table~\ref{tab:MetaLF_image_training} and global gradient-norm clipping at $1$. We use Adam parameters $(\beta_1,\beta_2)=(0.9,0.999)$ and $\epsilon=10^{-8}$, constant learning rates, and no weight decay. For ImageNet, we accumulate gradients over eight microbatches of eight images to obtain an effective batch size of $64$.

\begin{table}[h]
\centering
\small
\setlength{\tabcolsep}{6pt}
\caption{MetaLF training settings for image reconstruction. Settings are shared by the $a^{\mathbb{R}^2}$ and $a^{\mathrm{SE}(2)}$ variants. The inner coordinate count is per adaptation step.}
\label{tab:MetaLF_image_training}
\begin{tabular}{lccc}
\toprule
 & CIFAR10 & CelebA & ImageNet-1K \\
\midrule
Latent tokens $N$ & $25$ & $36$ & $169$ \\
Initial latent grid & $5\times5$ & $6\times6$ & $13\times13$ \\
$(K_{\mathrm{SA}},K_{\mathrm{CA}})$ & $(5,4)$ & $(6,4)$ & $(13,4)$ \\
Query/value frequency scales & $(1,3)$ & $(1,3)$ & $(2,10)$ \\
Inner / outer coordinates
 & $1{,}024 / 1{,}024$
 & $4{,}096 / 4{,}096$
 & $4{,}096 / 16{,}384$ \\
Effective batch size & $32$ & $32$ & $64$ \\
Training epochs & $150$ & $50$ & $2$ \\
Outer learning rate & $10^{-4}$ & $10^{-4}$ & $3\times10^{-4}$ \\
\bottomrule
\end{tabular}
\end{table}

\paragraph{Data and coordinate sampling}
CIFAR10 images retain their native $32\times32$ resolution. CelebA images are center-cropped to $140\times140$ and resized to $64\times64$ using bilinear interpolation with antialiasing. For both datasets, every pixel is used in each inner update and in the outer reconstruction loss. ImageNet training instead uses images at their native resolution, without resizing or cropping. Due to hardware limitations, we randomly sample $4{,}096$ pixels for each inner update and an independent set of $16{,}384$ pixels for the outer loss as also performed by~\citep{friedrich2025medfuncta}. The outer set may overlap the inner observations. Pixel-center coordinates are normalized independently along each axis to $[-1,1]$.

\paragraph{Model selection and evaluation}
For CIFAR10, we divide the original training set into $49{,}900$ training images and $100$ validation images. For CelebA, we use the official split of $162{,}770$ training, $19{,}867$ validation, and $19{,}962$ test images, with the first eight validation images used for checkpoint selection. For ImageNet, we hold out five training images per class, leaving $1{,}276{,}167$ training images and $5{,}000$ development images. In each case, we select the checkpoint with the highest mean validation PSNR. Final evaluation uses the $10{,}000$ CIFAR10 test images, the $19{,}962$ CelebA test images, and the $50{,}000$ official ImageNet validation images reserved as a test set. ImageNet evaluation resizes each image directly to $128\times128$ using antialiased bilinear interpolation and uses all $16{,}384$ pixels at each of the three inner steps. All datasets are scored on the full fitted grid.

\paragraph{Baselines} 
We re-run Functa and ENF reconstruction baselines that are also reported by~\citep{wessels2025grounding}. In general, we retrieve slightly higher test-set PSNR values among baselines, which may be due to slight differences in parameterization and a slightly longer optimization time. We notably identify much higher PSNR values for Functa on the CelebA and ImageNet-1K datasets, while obtaining a slightly lower PSNR for CIFAR10 (38.1dB compared to our 36.2dB). Functa is retrained with the settings proposed by~\citep{bauer2023spatial}. For consistency with ENF, we further adopt its latent parameterization, using the same number of latent tokens and content channels per token for each dataset.

\subsubsection{Shape reconstruction}
\label{app:shape_recon}
For shape reconstruction, a per-object SDF represents each ShapeNet sample, while each ACDC frame is represented by three jointly decoded SDF channels for the RV, LV blood pool, and LV myocardium. We initialize MetaLF with $N_{\mathrm{SA}}=2$ self-attention layers, hidden width $128$, four heads, and $32$ content channels per token, and we initialize query and value frequency scales to 1 and 3 respectively. Table~\ref{tab:shape_training} summarizes the remaining settings.

\begin{table}[h]
\centering
\small
\caption{Shape-reconstruction settings. Coordinate counts refer to
the SDF data term; ACDC additionally uses surface anchors.
ShapeNet settings are shared by the 16-category and 55-category
experiments.}
\label{tab:shape_training}
\begin{tabular}{lcc}
\toprule
 & ShapeNet & ACDC \\
\midrule
Latent tokens $N$ & $27$ & $256$ \\
Initial latent grid & $3\times3\times3$ & $8\times8\times4$ \\
$(K_{\mathrm{SA}},K_{\mathrm{CA}})$ & $(3,8)$ & $(6, 8)$ \\
Near-surface fraction & $75\%$ & $50\%$ \\
Inner / outer coordinates & $1{,}638/8{,}192$ & $24{,}000/24{,}000$ \\
Training epochs & $100$ & $200$ \\
\bottomrule
\end{tabular}
\end{table}

\paragraph{SDF preprocessing}
For ShapeNet, we construct an SDF cache from ShapeNetCore-v2
meshes as per~\citep{wessels2025grounding}. Each mesh is made watertight using Point Cloud Utils~\citep{point-cloud-utils} with reconstruction resolution $20{,}000$, centered at its
surface-area-weighted centroid, and scaled such that the longest bounding-box side has length one. We save $100{,}000$ uniform points in $[-0.5,0.5]^3$ and $100{,}000$ area-sampled surface points perturbed by isotropic Gaussian noise with standard deviation $0.01$, together with their signed distances.

For ACDC, we use the labelled end-diastolic and end-systolic segmentations. Each structure mask is zero-padded, converted to a surface by marching cubes at level $0.5$, and smoothed with $20$ Taubin iterations before computing signed distances. Each volume axis is normalized independently to $[-1,1]$. We cache uniform queries, near-surface queries with Gaussian perturbation scales $0.02$ and $0.005$, and exact surface points with their structure-channel identities. The pools contain $200{,}000$ uniform points, approximately $200{,}000$ near-surface points, and $100{,}000$ surface points per frame, with the latter two divided across the three structures.

\paragraph{Optimization and sampling}
All methods use full second-order MAML with $K=5$ latent updates. Adam~\citep{kingma2014adam} optimizes the decoder with learning rate $10^{-4}$ and global gradient-norm clipping at $1$. Meta-SGD~\citep{li2017meta} learns one rate per latent coordinate, shared across all five inner steps, using a separate Adam optimizer at $10^{-4}$ and clamping the rates to $[10^{-6},10]$. Rates are initialized to $0.05$ for posed coordinates and $0.01$ for content, while the Functa variants initialize all rates to $0.01$. Both optimizers use constant learning rates, $(\beta_1,\beta_2)=(0.9,0.999)$, and $\epsilon=10^{-8}$, without weight decay.

For each signal, we randomly sample coordinates as described in Table~\ref{tab:shape_training} without replacement, and randomly alternate entries to inner and outer sets. For ACDC, all $24{,}000$ inner observations are reused at every step. We additionally sample $1{,}024$ exact surface anchors per ACDC frame, shared by the inner and outer objectives.

\paragraph{Reconstruction objectives}
For ShapeNet, both objectives use the mean absolute difference between $\tanh(f_\theta(\mathbf{x};z)/\delta)$ and $\tanh(s(\mathbf{x})/\delta)$, with $\delta=0.1$ and target SDF $s$. No additional geometric penalties are used. For ACDC, the data term is instead the mean absolute difference between predicted and target SDFs clipped to $[-\delta,\delta]$. Both inner and outer objectives add an on-surface penalty $|f_c(\mathbf{x};z)|$ with weight $1$, an eikonal penalty $(\|\nabla_{\mathbf{x}}f_c\|_2-1)^2$~\citep{gropp2020implicit} with weight $0.1$, and a balanced inside/outside logistic loss with weight $1$.

\paragraph{Data splits and evaluation}
The ShapeNet-Part experiment uses ShapeNetCore-v2 instances
from the 16 ShapeNet-Part categories, while ShapeNet-Core uses all 55 available categories. We use category-stratified $80/10/10$ splits, yielding $22{,}456/2{,}800/2{,}822$ and $41{,}951/5{,}220/5{,}299$ training/validation/test shapes, respectively. IoU is evaluated on all $100{,}000$ cached uniform points per shape.

Of the pooled 150 patients from the ACDC dataset, we assign 120 patients to training, five to checkpoint selection, and 25 to testing. With two cardiac phases per patient, this gives 240 training, 10 selection, and 50 test volumes. At test time, we fit with SDF targets and the same composite objective used at training time. Predictions are scored against the original segmentation on the native voxel grid. All reported shape-reconstruction results report IoU percentages as mean and standard deviation averaged over three differently seeded training runs.

\paragraph{Baselines}
We train Functa, SpatialFuncta, and ENF using the same dataset-specific splits, sampling, objectives, and training budgets. ENF uses the same token counts and $32$ content channels as MetaLF, with hidden width $128$, attention dimension $64$, three heads, and eight cross-attention neighbours. SpatialFuncta uses the corresponding $3\times3\times3$ or $8\times8\times4$ grid with $32$ channels, trilinear interpolation, and a six-layer SIREN of width $256$ with $\omega_0=10$. This setting replaces the initial 15-layer, $\omega_0=30$ configuration, which failed to fit ShapeNet and became unstable during training on ACDC. Global Functa uses a $1{,}024$-dimensional latent and a 15-layer SIREN of width $256$ with $\omega_0=30$, for which its low scores reflect poor fitting under this five-step recipe. The content budgets of ENF, SpatialFuncta, and MetaLF are matched.

\subsubsection{Meta-classification}
For Table~\ref{tab:meta_cls}, we jointly train the representation and classifier through $K=3$ latent adaptation steps using full second-order MAML. The inner loop uses reconstruction supervision only, while class labels enter through the outer objective. We meta-learn the decoder, shared latent initialization, and classifier, for which the outer objective becomes
\begin{equation}
\mathcal{L}_{\mathrm{out}}=
\frac{\mathcal{L}_{\mathrm{rec}}}
     {\operatorname{sg}(\mathcal{L}_{\mathrm{rec}})}+
\frac{\mathcal{L}_{\mathrm{CE}}}
     {\operatorname{sg}(\mathcal{L}_{\mathrm{CE}})},
\end{equation}
where $\operatorname{sg}$ denotes stop-gradient and both losses are averaged over the current batch.

\paragraph{MetaLF configuration}
The MetaLF classifier mean-pools the final self-attention features, and subsequently applies a linear layer to return the class logits. Table~\ref{tab:metacls_training} lists any dataset-specific settings.

\paragraph{Optimization and augmentation}
Adam~\citep{kingma2014adam} updates the decoder and latent initialization at $3\times10^{-4}$, while a separate AdamW optimizer~\citep{loshchilov2017decoupled} trains the classifier with peak learning rate $2\times10^{-3}$, weight decay $0.05$, and cosine decay~\citep{loshchilov2016sgdr}. The classification head warmup is five epochs for CIFAR10 and ShapeNet and $249{,}500$ training images for ImageNet. During training, we utilize a random latent dropout rate of $60\%$ for the classification head. We further augment training with random horizontal flips.

\paragraph{Sampling and evaluation}
For CIFAR10, all methods sample all $1{,}024$ pixel grid locations during training and test-time. ImageNet training uses native-resolution images, reserving five training images per class for performance selection. A random sampler returns three inner sets of $2{,}048$ pixels and an independently sampled outer set of $8{,}192$ pixels. Final top-1 accuracy is reported on all $50{,}000$ official validation images. Accuracy uses native-resolution sampled adaptation with the same coordinate budget. The reported PSNR is calculated using images resized to $128\times128$ as in the reconstruction-only scenario (see Appendix~\ref{app:reconstruction_details}). ShapeNet-Core is similarly preprocessed as described in Appendix~\ref{app:shape_recon}. We report mean IoU and its sample standard deviation across test shapes. ShapeNet accuracy is the unweighted mean of the 55 category accuracies, with sample standard deviation across categories. Image accuracy uses the binomial standard error, and image PSNR variability is across test images.

\begin{table}[t]
\centering
\small
\caption{MetaLF meta-classification settings. Coordinate counts describe training; evaluation differences are specified below.}
\label{tab:metacls_training}
\begin{tabular}{lccc}
\toprule
 & CIFAR-10 & ImageNet-1K & ShapeNet-Core \\
\midrule
Latent tokens $N$ & $64$ & $256$ & $27$ \\
Initial latent grid & $8\times8$ & $16\times16$ & $3\times3\times3$ \\
Content channels per token & $16$ & $32$ & $32$ \\
Hidden width & $128$ & $256$ & $128$ \\
$N_{\mathrm{SA}}$ & $4$ & $4$ & $2$ \\
$(K_{\mathrm{SA}},K_{\mathrm{CA}})$ & $(8,4)$ & $(16,4)$ & $(6,8)$ \\
Inner / outer coordinates & $1{,}024/1{,}024$ & $2{,}048/8{,}192$ & $2{,}730/8{,}192$ \\
Training epochs & $100$ & $10$ & $100$ \\
Initial pose/content step sizes & $0.1/0.2$ & $0.1/0.2$ & $0.05/0.01$ \\
Meta-SGD optimizer learning rate & $0.2$ & $0.2$ & $10^{-4}$ \\
\bottomrule
\end{tabular}
\end{table}

\paragraph{Baselines}
We train Functa, SpatialFuncta, and ENF locally with the same objective function and dataset splits listed above. ENF and SpatialFuncta match MetaLF's token counts and content channels, while global Functa uses a $1{,}024$-dimensional latent on every dataset. All methods use mean pooling, layer normalization, and a linear classifier, reading MetaLF's final self-attention features, ENF's adapted content tokens, SpatialFuncta's grid modulations, or Functa's single global modulation vector.

\paragraph{Findings}
\label{app:cls_discussion}
As can be observed from Table~\ref{tab:meta_cls}, reconstruction quality for meta-classification is typically lower compared to when we only consider reconstruction quality as the outer objective (Table~\ref{tab:image_recon}). This is in line with earlier work on combined reconstruction and classification (MWT,~\citep{gielisse2025end}), indicating that reconstruction and classification are not necessarily synergistic tasks. This may be due to the potentially high in-class visual variability, which may result in distal latent reconstruction regions that are semantically close, requiring the model to make a trade-off.

Despite this trade-off, MetaLF extends naturally to meta-classification by adding a linear classification head and a classification loss to the outer objective, while retaining reconstruction-only adaptation in the inner loop. With random horizontal flips as the only image augmentation, MetaLF achieves 42.6\% top-1 accuracy on ImageNet-1K, exceeding the 24.1\% reported for MWT by 18.5 percentage points (Table~\ref{tab:meta_cls}). Development accuracy continued to improve towards the end of the 10-epoch training schedule, with no clear plateau, suggesting that longer training or a larger decoder may yield further gains.

On CIFAR10, MetaLF matches or exceeds the 82.1\% accuracy reported for a two-step ENF classifier, which trains a downstream P$\Theta$NITA classifier~\citep{bekkers2024fast} on representations meta-learned for reconstruction. MetaLF achieves an accuracy of 83.3\% using only random horizontal flips, compared to ENF using 50 random crops with flips per image. These results support the effectiveness of end-to-end MetaLF meta-classification with substantially less data augmentation.

\subsubsection{Meta-segmentation}
\label{app:meta_seg}
For Table~\ref{tab:meta_seg}, we jointly meta-learn reconstruction and segmentation using full second-order MAML. The decoder predicts signal values and segmentation logits as separate output channels of the same network. The inner loop adapts only the latent representation using reconstruction supervision, while the segmentation loss is only backpropagated in the outer objective. At test time, the decoder and learned step sizes remain fixed, and latent adaptation requires no segmentation labels.

For each batch, we optimize
\begin{equation}
\mathcal{L}_{\mathrm{out}} =
\frac{\mathcal{L}_{\mathrm{rec}}}
     {\operatorname{sg}(\mathcal{L}_{\mathrm{rec}})+10^{-8}}
+ \lambda(t)
\frac{\mathcal{L}_{\mathrm{seg}}}
     {\operatorname{sg}(\mathcal{L}_{\mathrm{seg}})+10^{-8}},
\end{equation}
where $\operatorname{sg}$ denotes stop-gradient. The segmentation term is binary cross-entropy for OMBRIA, 50-way cross-entropy for ShapeNet-Part, and 36-way focal loss with $\gamma=2$ for OASIS as also in~\citep{vyas2025fit}. We set $\lambda=1$ for OMBRIA and increase it linearly from zero to one over the first 50 training epochs for the two 3D datasets, updating the weight at every outer step.

\paragraph{Architecture and optimization.}
MetaLF uses four attention heads, while remaining settings are given in Table~\ref{tab:metaseg_settings}. Optimization is performed with AdamW~\citep{loshchilov2017decoupled}, which jointly updates the decoder and shared latent initialization with learning rate $3\times10^{-4}$, weight decay $10^{-4}$, and global gradient-norm clipping at $1$. A separate Adam optimizer~\citep{kingma2014adam} learns one Meta-SGD~\citep{li2017meta} step size per latent coordinate and inner step; its gradients are not clipped. For OASIS, the Meta-SGD optimizer rate is held at $0.2$ through epoch 10, decreased linearly to $0.05$ by epoch 20, and held constant thereafter. No data augmentation is used in these experiments.

\begin{table}[t]
\centering
\small
\caption{MetaLF meta-segmentation settings. Coordinate counts describe one training signal; ShapeNet-Part additionally supervises annotated surface points. The same number of inner steps is used during training and testing.}
\label{tab:metaseg_settings}
\begin{tabular}{lccc}
\toprule
 & OMBRIA & ShapeNet-Part & OASIS \\
\midrule
Latent tokens $N$ & $169$ & $27$ & $3510$ \\
Initial latent grid & $13\times13$ & $3\times3\times3$ & $13\times15\times18$ \\
Content channels per token & $64$ & $32$ & $64$ \\
Hidden width & $128$ & $128$ & $192$ \\
$N_{\mathrm{SA}}$ & $4$ & $2$ & $4$ \\
$(K_{\mathrm{SA}},K_{\mathrm{CA}})$ & $(13,4)$ & $(6,8)$ & $(13,8)$ \\
Query/value frequency scales & $(2,10)$ & $(1,3)$ & $(2,10)$ \\
Inner steps $K$ & $3$ & $5$ & $5$ \\
Inner coordinates & $2{,}048$ & $2{,}048$ & $32{,}768$ \\
Outer reconstruction coordinates & $8{,}192$ & $10{,}240$ & $32{,}768$ \\
Outer segmentation coordinates & $8{,}192$ & $2{,}048$ & $32{,}768$ \\
Batch size & $1$ & $4$ & $1$ \\
Training epochs & $200$ & $100$ & $200$ \\
Initial pose/content step sizes & $0.1/0.2$ & $0.05/0.01$ & $0.1/0.2$ \\
Meta-SGD optimizer rate & $0.2$ & $10^{-4}$ & $0.2\rightarrow0.05$ \\
\bottomrule
\end{tabular}
\end{table}

\paragraph{OMBRIA preprocessing and sampling}
Each $256\times256$ OMBRIA sample consists of co-registered Sentinel-1 and Sentinel-2 images before and after a flood. Sentinel-1 has one channel per time point and Sentinel-2 is an RGB image, giving eight reconstruction channels and one flood logit. Data is further split into $614/10/70$ training/validation/test examples.

For each outer loop, we draw $6{,}144$ coordinates which are divided into three $2{,}048$ coordinate inner sets, where each inner loop minimizes the eight-channel MSE. The outer set contains $4{,}096$ flood and $4{,}096$ background pixels, supervising both reconstruction MSE and binary cross-entropy simultaneously. Table~\ref{tab:meta_seg} pools intersections and unions over all 70 test images before taking their ratio.

\paragraph{ShapeNet-Part preprocessing and sampling}
For ShapeNet-Part segmentation, we join the official 2,048-point part annotations to the preprocessed ShapeNet data. SDF preprocessing follows the shape-reconstruction setup described in Section~\ref{app:shape_recon}. This gives a training/validation/test split of $12{,}082/1{,}866/2{,}868$ objects. Each outer loop iteration draws $20{,}480$ SDF queries within the uniform and near-surface coordinate pools, with the $25/75$ ratio described earlier. Both reconstruction objectives minimize mean absolute error after applying $\tanh(s/0.1)$ to predicted and target SDFs.

The decoder outputs one SDF value and 50 part class logits. Segmentation supervision is applied only at the 2,048 registered annotated surface points, using cross-entropy over all 50 parts. We first average IoU calculation over category parts within each object and subsequently over test objects, while test-time fitting uses the same SDF sampling budget with all annotated points getting scored. In Table~\ref{tab:meta_seg}, parentheses report the population standard deviation of instance mIoU over the 2,868 test objects.

\paragraph{OASIS preprocessing and sampling}
We use the Neurite OASIS-1 derivatives, which are affinely aligned, skull-stripped, bias-corrected T1 MRI with corresponding 35-structure segmentation, for which we assign $334/40/40$ of the 414 subjects to training/validation/testing. We center-crop the supplied $160\times192\times224$ volumes to $160\times160\times200$ as in~\citep{vyas2025fit}. The decoder predicts MRI intensity and 36 logits, including background.

Training operates on the cropped full-resolution volumes. Each of five inner steps receives $32{,}768$ observations, with $75\%$ sampled uniformly and $25\%$ proportional to MRI gradient magnitude. The inner loss is MSE divided by the stop-gradient mean squared target intensity of that step. The outer set has $32{,}768$ coordinates: $12.5\%$ uniform background, $12.5\%$ background adjacent to foreground, $25\%$ foreground voxels at interfaces between distinct foreground labels, and $50\%$ balanced over foreground labels. The outer objective uses uniformly weighted MRI MSE and a focal loss~\citep{lin2017focal}.

\paragraph{Local baselines and failed recipes}
Functa, SpatialFuncta, and ENF are trained with the same output-channel additions, reconstruction-only inner loop, joint outer objective, splits, and dataset-specific coordinate budgets. SpatialFuncta uses the MetaLF grid shape and content-channel count, with a six-layer, width-$256$ SIREN and $\omega_0=10$. Interpolation is bilinear for OMBRIA and trilinear for the 3D datasets. Its outer learning rate is reduced to $10^{-4}$ for OMBRIA and ShapeNet-Part, and remains $3\times10^{-4}$ for OASIS. ENF matches the token counts and content channels, uses attention dimension $64$, hidden width $128$ with three heads for OMBRIA and ShapeNet-Part, and width $192$ with four heads for OASIS. Its relative attributes, frequency scales, and cross-attention neighbour counts match MetaLF.

\begin{figure*}[t]
\centering
     \includegraphics[width=\textwidth, trim=0 0.4cm 0 0]{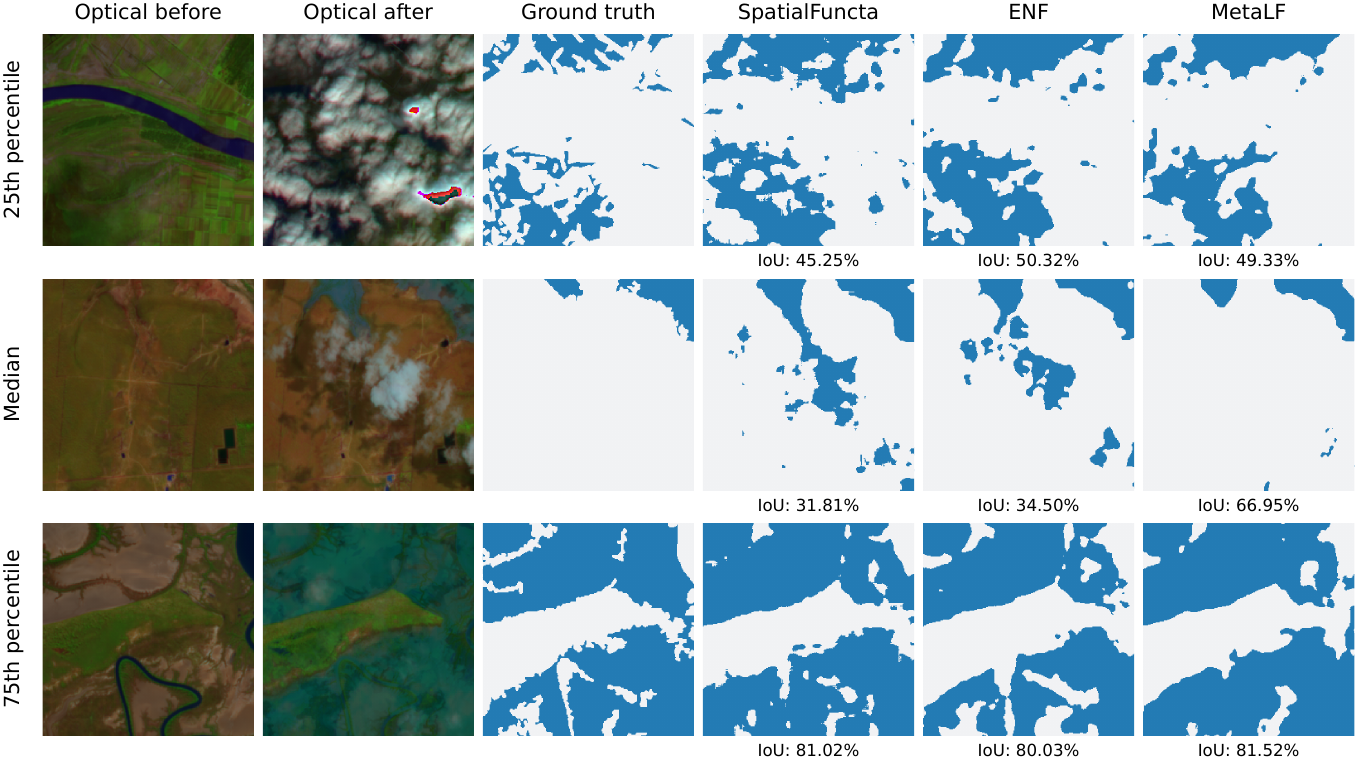}
     \caption{Qualitative example for OMBRIA meta-segmentation, illustrating the 25th percentile, median, and 75th percentile MetaLF test-set performance sample in terms of IoU. In the median case, SpatialFuncta and ENF produce extensive false-positive flood predictions around cloudy regions, whereas MetaLF largely suppresses them. This pattern is consistent with non-local latent interactions using broader spatial context to resolve ambiguous local signals.}
     \label{fig:ombria_segmentation}
\end{figure*}

Functa is standardized to use a 1,024-dimensional latent. Its OMBRIA result uses a six-layer, width-$256$, $\omega_0=10$ SIREN for which training collapsed mid-training. SpatialFuncta's ShapeNet-Part result likewise diverged at a late training stage. Thus, these two numeric entries do not represent completed 200- and 100-epoch runs, respectively. Global Functa failed to fit or diverged under the attempted ShapeNet-Part and OASIS recipes; these outcomes are reported as failed. The ShapeNet-Part attempts started from a 15-layer, width-$256$, $\omega_0=30$ SIREN and included seven single-setting interventions, none yielding a fitted model. The final OASIS attempt used a six-layer, width-$256$, $\omega_0=10$ SIREN and the common five-step recipe; it failed to fit before non-finite gradients stopped training at step $12{,}542$.

\paragraph{Findings} We note that MetaLF often performs similarly to spatially structured baselines (SpatialFuncta and ENF) when local information is sufficient for accurate segmentation. Our qualitative comparisons suggest that MetaLF distinguishes itself when long-range spatial context is needed to resolve ambiguous local cues. In the median OMBRIA example (see Figure~\ref{fig:ombria_segmentation}), SpatialFuncta and ENF produce extensive false-positive flood predictions around cloudy regions, whereas MetaLF largely suppresses these errors. This behaviour is consistent with MetaLF's non-local latent interactions allowing evidence from spatially separated regions to inform local predictions.

\subsection{Additional results}
\label{app:additional_results}

\subsubsection{Parameterization and computational costs}
\label{app:latent_ablation}

\paragraph{Latent pointcloud parameterization}

\begin{figure*}[h]
\centering
     \includegraphics[width=\textwidth, trim=0 0.2cm 0 0]{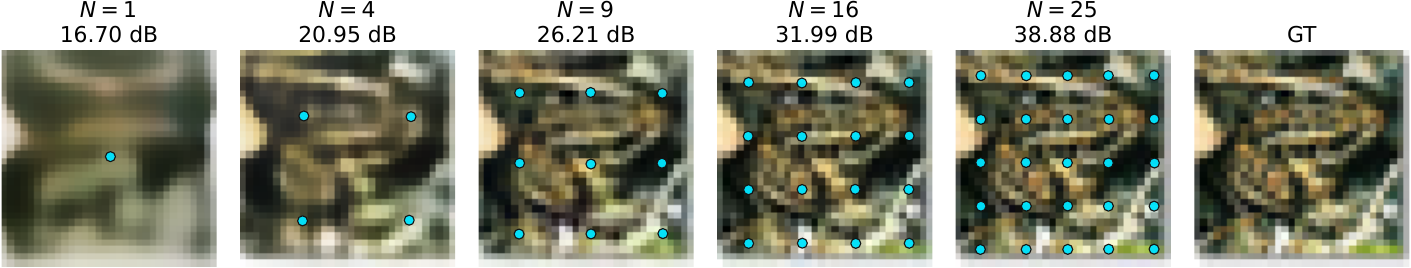}
     \caption{Reconstruction quality as a function of latent pointcloud size. The example uses MetaLF with $N_{\mathrm{SA}}=4$ and per-latent content size of $D=64$.}
     \label{fig:frog_latent_size}
\end{figure*}

We ablate latent pointcloud settings for CIFAR10 reconstruction in Table~\ref{tab:cifar-MetaLF-capacity}. With increasing pointcloud size, dimensionality, and number of MetaLF self-attention layers, we intuitively identify a steady improvement in reconstruction quality. Even for the diverse task of CIFAR10 reconstruction for which reconstruction can be considered a relatively localized task, we notice that the contextualizing expressivity of the MetaLF decoder can offer significant gains in reconstruction quality. Conversely, additional self-attention layers yield only small reconstruction gains when the number of latents $N=1$, consistent with the absence of interactions between distinct latent tokens. An example of how progressively increasing the granularity of the latent pointcloud $z$ affects reconstruction is shown in Figure~\ref{fig:frog_latent_size}.

\begin{table}[h]
\centering
\small
\caption{CIFAR-10 validation set reconstruction PSNR for MetaLF capacity variations. Rows vary the number of latents $N$; grouped columns vary latent channels $D$ and self-attention depth $N_{\mathrm{SA}}$.}
\label{tab:cifar-MetaLF-capacity}
\begin{tabular}{lcccccc}
\toprule
& \multicolumn{2}{c}{$D=16$} & \multicolumn{2}{c}{$D=32$} & \multicolumn{2}{c}{$D=64$} \\
\cmidrule(lr){2-3}\cmidrule(lr){4-5}\cmidrule(lr){6-7}
$N$ & $N_{\mathrm{SA}}=0$ & $N_{\mathrm{SA}}=4$ & $N_{\mathrm{SA}}=0$ & $N_{\mathrm{SA}}=4$ & $N_{\mathrm{SA}}=0$ & $N_{\mathrm{SA}}=4$ \\
\midrule
1 & 18.92 & 19.06 & 20.57 & 20.62 & 22.36 & 22.49 \\
4 & 22.69 & 23.03 & 24.77 & 25.29 & 27.58 & 28.42 \\
9 & 25.59 & 26.47 & 28.61 & 29.77 & 32.72 & 34.20 \\
16 & 28.47 & 29.85 & 32.54 & 34.17 & 39.17 & 40.78 \\
25 & 31.29 & 33.18 & 36.67 & 38.82 & 46.44 & \textbf{48.28} \\
\bottomrule
\end{tabular}
\end{table}

\paragraph{Computational cost analysis}
Table~\ref{tab:cifar10-compute} compares parameter counts, reconstruction costs, and meta-training costs on CIFAR10 at matched latent content budgets. Spatial Functa has the lowest costs among the evaluated configurations. Within MetaLF, additional self-attention layers increase parameter count and latency, while the increase in FLOPs is comparatively modest. ENF illustrates why fewer parameters do not necessarily imply less computation: its wide query-dependent transformations are repeatedly evaluated across the image. 

% Requires \usepackage{booktabs}.
\begin{table}[b]
\centering
\small
\setlength{\tabcolsep}{4pt}
\caption{Compute on CIFAR10 ($32\times32$) with three full-image inner loop steps and 1,600 content scalars, with ENF and MetaLF additionally fitting 50 pose scalars. Spatial Functa uses $5\times5\times64$ latents. Parameters include the shared decoder, trainable initialization, and trainable adaptation rates. Reconstruction includes encoding and decoding, while training includes a full second-order meta-update. Timings use one NVIDIA H100 GPU and report the median with the interquartile interval below, across 100 synchronized calls. FLOPs count logical XLA arithmetic (multiply-add $=2$), excluding transcendentals. Bold and underline denote the lowest and second-lowest costs.}
\label{tab:cifar10-compute}
\begin{tabular}{lccccc}
\toprule
& & \multicolumn{2}{c}{Reconstruction ($B=1$)}
    & \multicolumn{2}{c}{Meta-training ($B=32$)} \\
\cmidrule(lr){3-4}\cmidrule(lr){5-6}
Model & Params (M) $\downarrow$
      & GFLOPs $\downarrow$ & ms/image $\downarrow$
      & GFLOPs $\downarrow$ & ms/update $\downarrow$ \\
\midrule
Functa
& 6.187 & 11.975
& \shortstack{1.383\\{\scriptsize [1.382, 1.390]}}
& 1147.983
& \shortstack{\underline{40.367}\\{\scriptsize [40.364, 40.419]}} \\
Spatial Functa
& \textbf{0.354} & \textbf{3.846}
& \shortstack{\textbf{0.572}\\{\scriptsize [0.570, 0.573]}}
& \textbf{368.079}
& \shortstack{\textbf{14.543}\\{\scriptsize [14.537, 14.557]}} \\
ENF
& 0.610 & 33.232
& \shortstack{1.870\\{\scriptsize [1.868, 1.870]}}
& 3188.983
& \shortstack{116.081\\{\scriptsize [116.032, 116.099]}} \\
\midrule
MetaLF ($N_{\mathrm{SA}}=0$)
& \underline{0.362} & \underline{10.748}
& \shortstack{\underline{1.219}\\{\scriptsize [1.218, 1.220]}}
& \underline{1031.587}
& \shortstack{50.863\\{\scriptsize [50.836, 50.993]}} \\
MetaLF ($N_{\mathrm{SA}}=2$)
& 1.120 & 11.435
& \shortstack{3.173\\{\scriptsize [3.170, 3.174]}}
& 1088.456
& \shortstack{58.189\\{\scriptsize [58.082, 58.206]}} \\
MetaLF ($N_{\mathrm{SA}}=4$)
& 1.878 & 12.123
& \shortstack{5.131\\{\scriptsize [5.128, 5.139]}}
& 1145.331
& \shortstack{66.946\\{\scriptsize [66.894, 67.184]}} \\
\bottomrule
\end{tabular}
\end{table}

\subsubsection{Generative modeling} 
\label{app:generative}
\begin{figure*}[t]
\centering
     \includegraphics[width=\textwidth, trim=0 0.2cm 0 0]{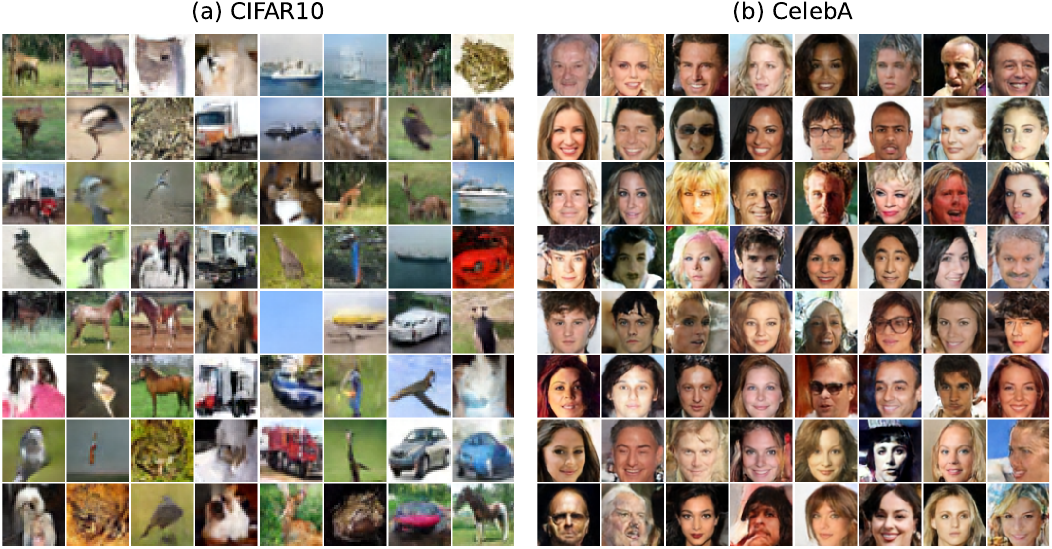}
     \caption{Uncurated examples generated using latent rectified flow based on the reconstruction meta-learned MetaLF (MetaLF-Diffuser) representation models.}
     \label{fig:generated_samples}
\end{figure*}

Though not the main focus of our work, we note that MAML implicitly provides structure to the latent space, as the limited number of optimization steps originating from the same initial latent forces together the latent representation in a contiguous and meaningful way~\citep{knigge2024space}. Therefore, we provide an additional experiment for generative modeling, where we compare ENF and MetaLF for which we train rectified flow~\citep{liu2022flow}, and for which results are presented in Table~\ref{tab:generative-modeling}. In both cases, we generate both parts of the latent pointcloud simultaneously, i.e. for \(
    z=\{(p_j,c_j)\}_{j=1}^{N}
\) we generate both points $p_j$ and content $c_j$ from random Gaussian noise. We can then decode the latents with their respective frozen, meta-learned neural field decoder (examples presented in Figure~\ref{fig:generated_samples}).

We evaluate generated images using Fr\'echet Inception Distance (FID)~\citep{chong2020effectively}, Kernel Inception Distance (KID)~\citep{binkowski2018demystifying}, and feature coverage. FID compares Gaussian approximations to real and generated Inception-feature distributions through their means and covariances, whereas KID measures their discrepancy using a polynomial-kernel maximum mean discrepancy. Lower values indicate closer distributional agreement. Coverage measures the fraction of real images with a generated neighbour within a feature-space neighbourhood defined by the fifth-nearest real neighbour. Higher coverage indicates broader coverage of the real-data distribution, complementing FID and KID.

On CelebA, MetaLF representations yield substantially better generative results than ENF under the same DiT backbone~\citep{peebles2023scalable}: FID decreases from $18.82$ to $13.05$, while coverage increases from $70.51\%$ to $86.43\%$. This is consistent with MetaLF being a more effective meta-learner for face representations: centered faces and the relatively consistent arrangement of facial features provide shared spatial regularities that its latent self-attention can exploit. On CIFAR10, where object identity, pose, background, and spatial arrangement vary more strongly between samples, the difference is smaller, with ENF and MetaLF performing similarly under DiT.

The MetaLF self-attention architecture also serves as a competitive diffusion-style generative backbone. Here, both backbones are trained with flow matching to predict a latent velocity field. Replacing DiT with the full-attention MetaLF-Diffuser (parameter-matched to DiT: 12 layers, 384 hidden width) improves CIFAR10 FID from $20.08$ to $18.83$ and coverage from $90.56\%$ to $92.87\%$, while retaining similar performance on CelebA (FID $13.38$ versus $13.05$), thus supporting utility beyond representation learning. Models are trained for 500 epochs.

\begin{table}[h]
\centering
\small
\caption{Generative modeling on CIFAR10 and CelebA using 50,000 generated samples, evaluated against 10,000 and 19,962 test images, respectively. CIFAR10 models are class-conditioned; CelebA models are unconditional. KID is reported as mean $\pm$ block standard error.}
\label{tab:generative-modeling}
\begin{tabular}{lcccccc}
\toprule
& \multicolumn{3}{c}{CIFAR10} & \multicolumn{3}{c}{CelebA} \\
\cmidrule(lr){2-4}\cmidrule(lr){5-7}
Model & FID $\downarrow$ & KID $\times 10^3$ $\downarrow$ & Cov.\ (\%) $\uparrow$
      & FID $\downarrow$ & KID $\times 10^3$ $\downarrow$ & Cov.\ (\%) $\uparrow$ \\
\midrule
ENF (DiT)
& \textbf{18.58} & \underline{6.89} $\pm 0.14$ & \textbf{93.06}
& 18.82 & $17.82 \pm 0.18$ & 70.51 \\
MetaLF (DiT)
& 20.08 & $7.22 \pm 0.18$ & 90.56
& \textbf{13.05} & $\mathbf{11.05} \pm 0.12$ & \underline{86.43} \\
MetaLF (MetaLF-Diffuser)
& \underline{18.83} & $\mathbf{6.59} \pm 0.19$ & \underline{92.87}
& \underline{13.38} & \underline{11.57} $\pm 0.14$ & \textbf{86.53} \\
\bottomrule
\end{tabular}
\end{table}

\end{document}